\documentclass[lettersize,journal]{IEEEtran}
\usepackage{colortbl}
\usepackage{multirow}
\usepackage{cite}
\usepackage{amsmath,amssymb,amsfonts}
\usepackage{graphicx}
\usepackage{textcomp}
\usepackage{array}
\usepackage{tabu}
\usepackage{xcolor}
\usepackage{subcaption}
\usepackage{balance}
\def\BibTeX{{\rm B\kern-.05em{\sc i\kern-.025em b}\kern-.08em
    T\kern-.1667em\lower.7ex\hbox{E}\kern-.125emX}}

\newcommand \includeall

\def\BibTeX{{\rm B\kern-.05em{\sc i\kern-.025em b}\kern-.08em
    T\kern-.1667em\lower.7ex\hbox{E}\kern-.125emX}}
\newcommand{\ApproxSign}{\raise.17ex\hbox{$\scriptstyle\sim$}}
\newcolumntype{C}[1]{>{\centering\let\newline\\\arraybackslash}m{#1}}
\usepackage{url}
\usepackage{caption} 
\onecolumn
\begin{document}
\title{WaferSegClassNet - A Light-weight Network for Classification and Segmentation of Semiconductor Wafer Defects}
\author{Subhrajit Nag, Dhruv Makwana, Sai~Chandra~Teja~R, Sparsh Mittal, C Krishna Mohan
\thanks{Subhrajit Nag is at IIT Hyderabad.  Dhruv Makwana and Sai~Chandra~Teja~R are at CKMVigil Pvt Ltd. Sparsh is at IIT Roorkee. C Krishna Mohan is at IIT Hyderabad and CKMVigil Pvt Ltd. Email of the corresponding author: saichandrateja@ckmvigil.in } }
%\author{}

%,~\IEEEmembership{Member,~IEEE}


%\markboth{}%
%{ : A Survey }

%\IEEEoverridecommandlockouts
% \IEEEpubid{\makebox[\columnwidth]{978-1-6654-4175-9/21/\$31.00~
% \copyright2021
% IEEE \hfill} \hspace{\columnsep}\makebox[\columnwidth]{ }}

\maketitle
 
%\IEEEcompsoctitleabstractindextext{%
%\IEEEtitleabstractindextext{%
\begin{abstract}
As the integration density and design intricacy of semiconductor wafers increase, the magnitude and complexity of defects in them are also on the rise. Since the manual inspection of wafer defects is costly, an automated artificial intelligence (AI) based computer-vision approach is highly desired. The previous works on defect analysis have several limitations, such as low accuracy and the need for separate models for classification and segmentation. For analyzing mixed-type defects, some previous works require separately training one model for each defect type, which is non-scalable. 

In this paper, we present WaferSegClassNet (WSCN), a novel network based on encoder-decoder architecture. WSCN performs simultaneous classification and segmentation of both single and mixed-type wafer defects. WSCN uses a ``shared encoder'' for classification, and segmentation, which allows training WSCN end-to-end. We use N-pair contrastive loss  to first pretrain the encoder and then use BCE-Dice loss for segmentation, and categorical cross-entropy loss for classification. Use of N-pair contrastive loss helps in better embedding representation in the latent dimension of wafer maps.  WSCN has a model size of only 0.51MB and performs only 0.2M FLOPS. Thus, it is much lighter than other state-of-the-art models. Also, it requires only 150 epochs for convergence, compared to 4,000 epochs needed by a previous work.  We evaluate our model on the MixedWM38 dataset, which has 38,015 images. WSCN achieves an average classification accuracy of 98.2\% and a dice coefficient of 0.9999. We are the first to show segmentation results on the MixedWM38 dataset. The source code can be obtained from \url{https://ckmvigil.github.io/wscn/}

%\url{https://sites.google.com/view/wafer-defect-classify-segment/}. 
  
\end{abstract}

% Note that keywords are not normally used for peer review papers.
% \begin{IEEEkeywords}
% Review, deep neural networks,  CPU-GPU heterogeneous computing, smartphone, distributed computing.   
% \end{IEEEkeywords}
%}

\maketitle
%\IEEEdisplaynotcompsoctitleabstractindextext
% \fi
\IEEEpeerreviewmaketitle

\ifCLASSOPTIONcompsoc
  \noindent\raisebox{2\baselineskip}[0pt][0pt]%
  {\parbox{\columnwidth}{\section{Introduction}\label{sec:introduction}%
  \global\everypar=\everypar}}%
  \vspace{-1\baselineskip}\vspace{-\parskip}\par
\else
  \section{Introduction}\label{sec:introduction}\par
\fi
% \IEEEPARstart{T}{his demo} file is   
 
Semiconductors are used in all modern electronic devices and technologies. Hence, semiconductor chips have become the key driver for industrial growth.  Given the intense quality and low-cost demands placed on semiconductor manufacturing, achieving a high yield (fraction of fault-free devices out of total devices) is essential to generate higher revenues. As a result, yield-based quality is the primary criterion for a company to succeed in the market. 

Wafer fabrication is a  costly and complex  process involving many process steps involving many process variables.  Finally, the wafers are evaluated through end-of-line (EOL) tests for ensuring the products perform the desired functionality.  One of these tests outputs a wafer map (WM) showing the count and locations of defective dies on every wafer. The increasing complexity of wafer design and increasing integration density have aggravated defects in semiconductor wafers. This has also led to mixed-type defects, where two or more defect patterns exist in a single wafer. Upon identifying the correct defect type, the engineers can localize the cause(s) of the defects in the process, correct them and increase the yield. Yet, classifying mixed-type defects is especially challenging since the location and angle of single defects can vary widely. Hence, they may combine in numerous ways to create mixed-type defects. Efficient detection of these defects is a crucial challenge faced by semiconductor manufacturing companies. 

Traditionally, wafer defect detection has been done manually by experienced engineers \cite{hansen1998use}.  However, this process is cumbersome and unscalable. Artificial intelligence (AI) has shown promising results in many real-world applications \cite{capizzi2018optimizing}. 
An AI-based computer-vision approach  \cite{ref129} allows defect detection in a contact-free manner. An AI model can extract edge features, surface textures, and pattern information to detect the defects accurately \cite{sciuto2019organic}. However, previous works on wafer defect detection have crucial limitations (refer Section \ref{sec:relatedWork} for more details). Some works have been evaluated on minuscule datasets \cite{saad2015defect,han2020polycrystalline} and others have tested for only a small number of defects \cite{nakazawa2019anomaly}. A few other works provide low classification accuracy (93.2\%) \cite{wang2020deformable}. All the previous works have only performed either classification or segmentation, but not both. Many previous works have mostly used synthetic WMs with simulated defects \cite{kyeong2018classification}. Thus, there is a need to design novel network for wafer defect classification and segmentation and evaluate them on large-scale realistic datasets. 

\textbf{Contributions:} In this paper, we present WaferSegClassNet (WSCN), a deep-learning model for simultaneously performing both classification and segmentation of defects on WMs. To the best of our knowledge, WSCN is the first wafer defect analysis model that performs both segmentation and classification. WSCN follows a multi-task learning framework to segment and classify an image simultaneously.  

WSCN follows encoder-decoder architecture. The decoder has two branches: one for segmentation and one for classification. WSCN is carefully designed to achieve high predictive performance yet remain a lightweight network. For example, the encoder uses separable convolutions to reduce the memory bandwidth and computational requirements. WSCN uses a ``shared encoder'' for segmentation and classification, which allows training WSCN end-to-end. WSCN has a model size of only 0.51MB and performs only 0.2M FLOPS. Thus, it is much lighter than other state-of-the-art models. By performing INT8 quantization, the model size of WSCN decreases further by three times, with negligible impact on predictive performance. 

For supervised learning, the most widely used loss function is the ``cross-entropy loss'', however, it has crucial limitations (refer Section \ref{sec:lossfunction} for more details). We use N-pair contrastive loss to first train the encoder for 100 epochs and then use BCE-DICE loss to train segmenation branch and Categorical Cross-entropy loss
for classification branch for 50 epochs. Use of N-pair contrastive loss helps in better embedding representation in latent dimension of WMs having single defect and mixed defect types. This improves classification and segmentation results. Use of N-pair contrastive loss also reduces the time to train. Specifically, WSCN requires only 150 epochs for convergence, compared to 4,000 epochs needed by the DCNet  \cite{wang2020deformable} model.

We evaluate WSCN on a large dataset  viz., the MixedWM38 dataset \cite{waferDefectDataset} (refer Section \ref{sec:MixedWM38} for more details). It has 38,015 images, corresponding to 38 single and mixed type defect patterns, comprehensively covering all the defect patterns. For defect classification, we compare our model with ResNet50 \cite{he2016deep}, DenseNet121 \cite{huang2017densely}, LeNet \cite{lecun1998gradient}, AlexNet \cite{krizhevsky2012imagenet}, EfficientNetB0 \cite{tan2019efficientnet}, MobileNetV2 \cite{sandler2018mobilenetv2}, ResNet18 \cite{he2016deep}, and DCNet \cite{wang2020deformable}, a recently proposed CNN for wafer defect classification. WSCN achieves an average classification accuracy of 98.20\%, which is much better than the 93.20\% accuracy achieved by DCNet. The ResNet50  and DenseNet121 models achieve an accuracy of 97.19\% and 98.34\%, respectively. However, these models have a much larger model size. In fact, the model size of ResNet50 is 200$\times$ that of WSCN. LeNet, ResNet18, AlexNet, EfficientNetB0 and MobileNetV2 also achieve lower classification accuracy than WSCN.
 
A key challenge in analyzing mixed-type defects is that one defect may overlap with another defect. Since object detection only provides a bounding box around the defect, it is insufficient for separately detecting different defects. Hence, we perform defect \textit{segmentation} and not defect \textit{detection}.
In MixedWM38 dataset, each WM has three regions: wafer boundary, background and defect. MixedWM38 dataset provides class labels, which are required for classification.   However, MixedWM38 dataset does not provide annotations for bounding box detections and segmentation masks. To get around this limitation, we combine wafer boundary and background into a single class, and the defect into another class. We then perform binary segmentation to segment the defects (Section \ref{sec:preprocessing}).

As for segmentation results, WSCN achieves a dice coefficient of 0.9999 and an IoU value of 0.9999. These values indicate the effective segmentation capability of WSCN. Since no previous work has performed segmentation on this dataset, we compare WSCN with open-source segmentation models such as UNet \cite{ronneberger2015u}, and DeepLabV3+ \cite{chen2017deeplab}. These models achieve a similar dice coefficient as WSCN. However, they require a much larger model size, e.g., the model size of UNet is 250$\times$ that of WSCN. The frame-rate of WSCN is 25.11 frames-per-second on the Tesla P100 GPU. This demonstrates that WSCN can work in real-time to perform defect detection with high-performance.
 
The remainder of this paper is organized as follows.  Section \ref{sec:relatedWork} discusses related works on wafer defect analysis. Section \ref{sec:MixedWM38} discusses the MixedWM38 dataset. Section \ref{sec:ProposedApproach} discussed our proposed model. Section \ref{sec:experimentalResults} presents the experimental platform, and comparative results.  Finally, Section \ref{sec:Conclusion}  concludes this work.

\section{Related work} \label{sec:relatedWork}

\textbf{Related works on defect classification:} 
Nakazawa et al. \cite{nakazawa2018wafer} use CNNs for wafer defect classification. They train and validate their CNN models using synthetically generated WM. The testing is done on generated as well as on real WMs. 
Maksim et al. \cite{maksim2019classification} evaluate the performance of well-known CNN models like MobileNetV2, VGG19, ResNet50, and ResNet34 on various WM defects. To address the problem of data scarcity, they make their own composite dataset consisting of synthetic data and experimental data. This composite dataset is used for training the models. Testing is done on the WM–811K dataset \cite{wu2014wafer}.
Wang et al. \cite{wang2020deformable} propose a ``deformable convolutional network (DCNet)'' for defect classification.  DCNet uses ``deformable convolution'' (DC) to extract the defects' feature representations by focusing on the sampling area on defective dies.  The output layer is multi-labeled and is one-hot encoded. This transforms the mixed type of defects into individual single defects and ensures effective identification of each defect. 

 Kyeong et al. \cite{kyeong2018classification} train separate classification models for classifying single defects and then use all those models for finding the presence of mixed-type defects. However, such an approach is not scalable because it significantly increases storage and computation overhead. Also, it requires separately training for all those models. Further, they study only four single defect patterns, namely ``circle'', ``ring'', ``scratch'', and ``zone''. By contrast, we propose a single model for classifying single-type and mixed-type defects. Also, we test our model on 8 single-type and 29 mixed-type defect patterns.
 
  Elliptical basis function neural networks use a hidden layer of elliptical units. They handle every input individually using separate weights. Sciuto et al. \cite{sciuto2019organic,lo2021organic} use elliptical basis neural networks for detecting defects in organic solar cells.

\textbf{Related works on defect segmentation:} 
Saad et al. \cite{saad2015defect} convert the wafer image from ``RGB-space'' to ``L*a*b*'' space, where all the color information is present in the a*b* spaces only. Then, they use k-means clustering to separate the pixels into two clusters, representing the defect and wafer, respectively. Nakazawa et al. \cite{nakazawa2019anomaly} use encoder-decoder-based models for detection and segmentation of abnormal patterns of wafer defects. They generate abnormal wafer defects using ``defect pattern generation model''. They develop three separate encoder-decoder models based on SegNet \cite{badrinarayanan2017segnet}, U-Net \cite{ronneberger2015u} and FCN \cite{long2015fully}. 
Table \ref{tab:comparison} compares selected previous works on key parameters.

\begin{table*}[htbp]
  \centering
  \caption{Comparison of related works}
    \begin{tabular}{|c|c|c|c|c|}
    \hline
    Work  & Objective & Types of defects & Network & Dataset \\
    \hline
    Nakazawa et al. \cite{nakazawa2018wafer} & Classification & 22 defects & CNN   & 28.6K for training, 1.19K for testing \\
    \hline
    Wang et al. \cite{wang2020deformable} & Classification & 8 single, 29 mixed & CNN based on deformable CONV & 38015 images \\
    \hline
    Saad et al. \cite{saad2015defect} & Segmentation & 1 & k-means clustering & 2 images \\
    \hline    
    Nakazawa et al. \cite{nakazawa2019anomaly} & Segmentation & 11 defects & FCN, SegNet and U-Net & 17K for training, 3.3K for testing \\
    \hline    
    Han et al. \cite{han2020polycrystalline} & Segmentation &  1 & RPN+U-net with dilated CONV & 106 images \\
    \hline
    Ours  & Classification+Segmentation & 8 single, 29 mixed & Encoder-decoder CNN & 38015 images \\
    \hline
    \end{tabular}%
  \label{tab:comparison}%
\end{table*}%

\section{The MixedWM38 Dataset}\label{sec:MixedWM38}
\subsection{Dataset description}
The MixedWM38 dataset \cite{waferDefectDataset} was  published by Wang et al. \cite{wang2020deformable}.  This dataset has 38 defect patterns, viz., one fault-free pattern, eight single defect patterns, thirteen 2 mixed-type patterns, twelve 3 mixed-type defect patterns, and four 4 mixed-type defects for a total of  38,015 WMs.  
The dataset contains images of $52\times52$ dimension and class information as one hot encoded array in numpy format.
Table \ref{tab:cause} describes these defects along with their possible causes. Table \ref{tab:defectpattern} shows 38 defect patterns.

\begin{table*}[htbp]
  \centering
  \caption{Single-type defect classes and their possible causes (WM =wafer map, acr. = acronym)}
    \begin{tabular}{|l|c|p{5cm}|p{10cm}|}
    \hline
    Name  & Acr. & Description & Cause of defect \\
    \hline
    Center & C     & Defective die scattered in the centre of WM & Abnormality of RF (Radio Frequency) power, abnormality in liquid flow or abnormality in liquid pressure. Problem in the plasma area \cite{hansen1998use}. Thin film deposition \cite{piao2018decision} or abnormality in liquid pressure. \\
    \hline
    Donut & D  & Defective die from centre of WM in a ring configuration   & Redeposition of dissolved photoresist solids backing onto wafer during developing process, as the center rinsing step results less defective at the center of the wafer. \\
    \hline
    Edge-Loc & EL & Localized clusters on the edge   &  Uneven heating during diffusion process \cite{hansen1997monitoring}. \\
    \hline
    
    Edge- Ring & ER & Ring-shaped clusters around  perimeter & Abnormal temperature control in the rapid thermal annealing process \cite{tello2018deep}. \\
    \hline
    Local type & L  & Localised clusters occuring regularly   & Excess vibration in a given machine liberating enough particles \cite{hansen1998use}. Crystalline heterogeneity \cite{jeong2008automatic}. Due to silt valve leak, abnormality during robot handos or abnormality in the pump. \\
    \hline
    Nearful & NF  & Unusual fault pattern    & Reception failure.    \\
    \hline
    Scratch & S & Distribution of faulty dies in a long, narrow region    & Human error at the shipping and handling process \cite{jeong2008automatic}. Error at chemical mechanical polishing (CMP).  \\
    \hline
    
    Random & R & Random defect without patterns  & Contaminated pipes, abnormality in shower-head, or abnormality in control wafers. \\
    \hline
    
    \end{tabular}%
  \label{tab:cause}%
\end{table*}%

\begin{table}[htbp]
  \centering
  \caption{{38 defect patterns in MixedWM38 dataset}}
   \begin{tabular}{|c|c|c|c|c|c|c|c|}    
    \hline
        No.  & Single defect  &   No.    & Two-mixed defect &  No.    & Three-mixed defect  &  No.     &  Four-mixed defect  \\
    \hline
    1     & Normal & 10    & C+EL  & 23    & C+EL+L & 35    & C+L+EL+S \\
    \hline
    2     & Center (C) & 11    & C+ER  & 24    & C+EL+S & 36    & C+L+ER+S \\
    \hline
    3     & Donut (D) & 12    & C+L   & 25    & C+ER+L & 37    & D+L+EL+S \\
    \hline
    4     & edgeLoc (EL) & 13    & C+S   & 26    & C+ER+S & 38    & D+L+ER+S \\
    \hline
    5     & edgeRing (ER) & 14    & D+EL  & 27    & C+L+S &       &  \\
    \hline
    6     & Loc (L) & 15    & D+L   & 28    & D+EL+L &       &  \\
    \hline
    7     & Nearful (NF) & 16    & ER+L  & 29    & D+EL+S &       &  \\
    \hline
    8     & Scratch (S) & 17    & EL + S & 30    & D+L+S &       &  \\
    \hline
    9     & Random (R) & 18    & ER+S  & 31    & D+ER+L &       &  \\
    \hline
          &       & 19    & L+S   & 32    & D+ER+S &       &  \\
    \hline
          &       & 20    & D+ER  & 33    & EL+L+S &       &  \\
    \hline
          &       & 21    & D+S   & 34    & ER+L+S &       &  \\
    \hline
          &       & 22    & EL+L  &       &       &       &  \\
    \hline    
    \end{tabular}%
  \label{tab:defectpattern}%
\end{table}%

\subsection{Data Preparation and Preprocessing}\label{sec:preprocessing}
The dataset is divided into training and validation set containing 80 and 20 percent of the dataset, which is 30412 and 7603 images, respectively. In the MixedWM38 dataset, each WM has three regions: wafer boundary, background and defect. MixedWM38 dataset provides class labels, which are required for classification.   However, MixedWM38 dataset does not provide annotations of segmentation masks. To get around this limitation, we combine wafer boundary and background into a single class and the defect into another class and then perform binary segmentation to segment the defects. We reshape the dataset images from $52\times52$ to $224\times224$ to learn features and patterns more effectively. To match the dimension of masks with the dimensions of the input image, we reshape masks to the shape of $224\times224$. We perform one-hot encoding on ground truth class labels to generate ground truth of shape $B\times1\times38$, which is required for WSCN model, where B is the batch size.

\section{Proposed Approach} \label{sec:ProposedApproach}
We present the overall architecture of our proposed  WaferSegClassNet network (Section \ref{sec:overallArchitecture}), the encoder and decoder stages (Section \ref{sec:encoder}-\ref{sec:decoderStage}). We also discuss the loss function used for pretraining the encoder (Section \ref{sec:lossfunction}).

\subsection{WaferSegClassNet architecture} \label{sec:overallArchitecture}
The diversity of defect shapes and defect overlap for semiconductor wafer defect datasets has posed new challenges for identifying WM defects. To solve these challenges and enable CNN to identify different types of defect shapes and defects overlap, we propose a novel model, WaferSegClassNet (WSCN), that can perform both defect classification and segmentation. WSCN model follows a multi-task learning framework that allows it to learn to segment and classify an image simultaneously. A classification network works with global information, whereas a segmentation network incorporates more local information in the input image. Thus, we propose a single network that focuses both on global and local information. Some previous works (e.g., \cite{wang2006detection}) remove global random defects before applying classification or clustering. Our CNN-based approach does not require such ad-hoc steps, nor does it require clustering.

\begin{figure}[htbp]  \centering
\includegraphics [scale=0.70] {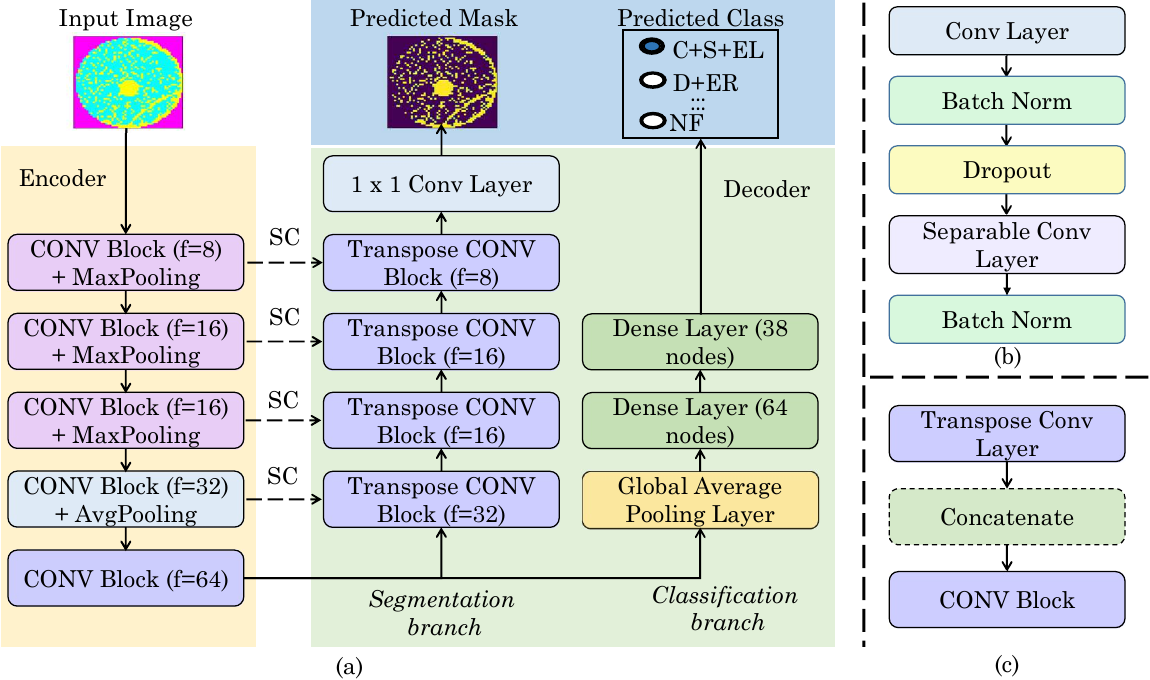}
\caption{a) WaferSegClassNet architecture (`f' shows the number of filters; SC = skip-connection), (b) CONV Block, (c) Transpose CONV block }
 \label{fig:ModelArchitecture}
  \end{figure}

Figure \ref{fig:ModelArchitecture}(a) shows the overall architecture of  WaferSegClassNet. Here, the spatial dimension of the input is 224 $\times$ 224, and the output dimension of the encoder branch is 14 $\times$ 14. WSCN follows encoder-decoder architecture, where the encoder performs downsampling and the decoder performs upsampling. In Figure \ref{fig:ModelArchitecture}(a), $f$ denotes the number of filters. At each downsampling layer, f is used as 8, 16, 16, 32, and 64, respectively. At each up-sampling layer, f is used as 32, 16, 16 and 8, respectively. We now describe the encoder and decoder stages of our model.

\subsection{Encoder stage}\label{sec:encoder}
The encoder uses a series of convolution blocks with pooling layers to extract multi-scale local details.  As depicted in Figure \ref{fig:ModelArchitecture}(b), the convolution block contains a 3 $\times$ 3  convolution layer with batch-normalization and dropout layers and a 3 $\times$ 3 separable convolution layer with batch-normalization. 
We use separable convolutions in the CONV block to reduce the memory bandwidth and computational requirements while also improving the representational efficiency. After the last convolution block, we use ``average pooling'' to linearly transform the vectorized feature maps. Addition of this average pooling layer acts as an effective regularizer and ensures no information loss. This is important for accurately classifying and segmenting the semiconductor wafer defects.

\subsection{Decoder stage}\label{sec:decoderStage}
In our network, the decoder block serves a dual purpose of performing classification and producing the semantic segmentation masks by recovering the spatial information.

\textbf{Segmentation branch:} The segmentation branch produces the binary segmentation mask. It has four transpose CONV blocks of $3\times3$  and a $1\times1$ convolution layer. The segmentation branch contains transpose CONV blocks for generating the output mask by upsampling the feature representation. As shown in Figure \ref{fig:ModelArchitecture}(c), transpose CONV blocks contain the transpose convolution layer.  Transpose convolution upsamples the ``input feature map'' to a desired ``output feature map'' using learnable parameters.
It interprets the coarse input data to fill-in the detail during upsampling. In the transpose convolution block, the concatenation (``concatenate'') layer is used to enrich the semantic information of input image through skip connections. This reduces the error rate of mask generation.

\textbf{Classification branch:} The classification branch identifies the defect type based on the features extracted by the encoder. The classification branch contains ``global average pooling'' (GAP) layer and two dense layers with 64 and 38 nodes, respectively. We have used GAP layer because it is more native to the convolution structure and it enforces correspondence between feature maps and categories. GAP layer has no learnable parameters and hence, it avoids overfitting. GAP layer makes the network more robust towards spatial translations due to its nature of adding spatial information. The last dense layer outputs 38 class probabilities corresponding to 38 defect types.

\subsection{N-pair contrastive loss function}\label{sec:lossfunction}
The most widely used loss function for supervised learning is the ``cross-entropy loss''. However, it has crucial limitations \cite{zhang2018generalized}, such as noisy labels causing lack of robustness and poor performance with adversarial examples. In this paper, we use N-pair contrastive loss  \cite{sohn2016improved} for supervised learning where normalized embeddings from the same category are drawn together more closely than embeddings from different categories.
This loss is used to train the encoder to  generate vector embeddings of input images so that representations of images in the same class category are more similar than representations in different classes. Experimentally, it outperforms supervised training with ``cross-entropy loss''. Training an encoder with N-pair contrastive loss helps better embedding representation in the latent dimension of semiconductor WM having single defect and mixed defect types. This improves the accuracy of the overall network. Pre-training an encoder with N-pair contrastive loss reduces the distance between similar embeddings, resulting in better feature learning in the encoder stage, which aids in both classification and segmentation branches of the decoder.

We pre-train the encoder for 100 epochs with N-pair contrastive loss to generate feature representation of input data. We  use the learned feature representations from the encoder block in the segmentation branch, and classification branch by freezing weights learned from the encoder branch. Instead of a ``dedicated network'' for segmentation and classification, a ``shared encoder'' makes it possible to train WSCN end-to-end, thereby reducing training time and model size. 
Also, a ``shared encoder'' takes advantage of the shared information across multiple tasks and thus, helps in improvement of the model performance for all tasks.

\section{Experimental Results}\label{sec:experimentalResults}
In this section, we describe the evaluation approach (Section \ref{sec:evaluation}), and present the results of classification (Section \ref{sec:classification}), including ROC-AUC curve (Section \ref{sec:ROC}). We also analyze the failure cases to gain more insights (Section \ref{sec:mispredictedCases}). We then present the results of segmentation (Section \ref{sec:segmentation}). Finally, we report the latency value, and the model size reduction achieved by quantization (Section \ref{sec:latency}) and the limitation of the proposed methodology (Section \ref{sec:limitation}). 

\subsection{Evaluation approach}\label{sec:evaluation}
  
\textbf{Experimental platform:} 
The experimental tests were conducted using TensorFlow-GPU 2.6, and CUDA 11.2.
The initial ``learning rate'' is 0.001 and the ``decay rate'' is 0.1 whenever there is no convergence for 10 continuous epochs. 
The batch size is 64.  We use Adam optimizer with N-pair contrastive loss for pre-training the encoder for 100 epochs. We then use Adam optimizer with BCE-Dice loss for training the segmentation network and categorical cross-entropy loss for training the classification for 50 epochs. BCE-DICE loss combines ``binary cross-entropy loss'', which is ``distribution-based loss''  with ``DICE loss'' which is ``region-based loss''.  

\textbf{Metrics:} For evaluating classification performance, we use accuracy, MCC \cite{chicco2020advantages}, ROC-AUC curve, precision and recall metrics. In the ROC curve, the ``area under curve'' (AUC) value ranges from 0 to 1. The higher the AUC, the better is the model's performance at distinguishing between the positive and negative classes. Matthews correlation coefficient (MCC) is defined as follows:

$MCC = \frac{TP * TN - FP * FN}{ \sqrt{ (TP + FP)(TP+FN)(TN+FP)(TN+FN)} } $

 A value of -1 and +1 indicate perfect misclassification and perfect classification, respectively. 
 MCC considers true/false positive/negative and hence, is a better metric than ROC-AUC curve. MCC is the only binary classification rate that a high score only if the binary classifier could correctly predict a majority of positive instances and the majority of negative instances  \cite{chicco2020advantages}. MCC is especially useful for mitigating the class imbalance issue  \cite{chicco2020advantages}. 
 
 For evaluating the performance of segmentation,  we use the dice coefficient, also known as dice similarity coefficient and intersection over union (IoU) \cite{MetricDefinition}. Although similar, these metrics have subtle differences \cite{DiceScoreIoU} and hence, we have included both of them.

\subsection{Classification Results}\label{sec:classification}

We compare the results of classification for our model WSCN with DCNet \cite{wang2020deformable}, ResNet50 (shown as ResNet) \cite{he2016deep} and DenseNet121 (shown as DenseNet) \cite{huang2017densely}. This is because DCNet was also trained on the same dataset and ResNet50 and DenseNet121 are well-known, open-source models that can be trained on any dataset.
Table \ref{tab:classwise} presents classwise accuracy, precision and recall results. 

\begin{table*}[htbp]
  \centering
  \caption{Classwise accuracy results. Avg(1Defect) shows average of eight single defect patterns. Avg(2Defects) shows average of thirteen 2 mixed-type patterns. Average(3Defects) shows average of twelve 3 mixed-type defect patterns. Average(4Defects) shows the average of  four 4 mixed-type defects.}
    \begin{tabular}{|c|r|r|r|r|r|r|r|r|r|r|r|r|}
    \hline
Class  & \multicolumn{4}{c|}{Accuracy} & \multicolumn{4}{c|}{Precision} & \multicolumn{4}{c|}{Recall} \\    \hline
    
    \multicolumn{1}{|c|}{} & \multicolumn{1}{c|}{DenseNet} & \multicolumn{1}{c|}{ResNet} & \multicolumn{1}{c|}{DCNet} & \multicolumn{1}{c|}{WSCN} & \multicolumn{1}{c|}{DenseNet} & \multicolumn{1}{c|}{ResNet} & \multicolumn{1}{c|}{DCNet} & \multicolumn{1}{c|}{WSCN} & \multicolumn{1}{c|}{DenseNet} & \multicolumn{1}{c|}{ResNet} & \multicolumn{1}{c|}{DCNet} & \multicolumn{1}{c|}{WSCN} \\
    \hline
    \multicolumn{1}{|c|}{1} & \cellcolor[rgb]{ .353,  .541,  .776}100.00\% & \cellcolor[rgb]{ .353,  .541,  .776}100.00\% & \cellcolor[rgb]{ .451,  .612,  .812}99.70\% & \cellcolor[rgb]{ .353,  .541,  .776}100.00\% & \cellcolor[rgb]{ .353,  .541,  .776}100.00\% & \cellcolor[rgb]{ .353,  .541,  .776}100.00\% & \cellcolor[rgb]{ .984,  .925,  .937}94.00\% & \cellcolor[rgb]{ .353,  .541,  .776}100.00\% & \cellcolor[rgb]{ .353,  .541,  .776}100.00\% & \cellcolor[rgb]{ .353,  .541,  .776}100.00\% & \cellcolor[rgb]{ .984,  .878,  .89}91.00\% & \cellcolor[rgb]{ .353,  .541,  .776}100.00\% \\
    \hline
    \multicolumn{1}{|c|}{2} & \cellcolor[rgb]{ .353,  .541,  .776}100.00\% & \cellcolor[rgb]{ .353,  .541,  .776}100.00\% & \cellcolor[rgb]{ .984,  .984,  .996}97.80\% & \cellcolor[rgb]{ .353,  .541,  .776}100.00\% & \cellcolor[rgb]{ .671,  .765,  .89}99.00\% & \cellcolor[rgb]{ .522,  .659,  .835}99.48\% & \cellcolor[rgb]{ .984,  .91,  .922}93.00\% & \cellcolor[rgb]{ .671,  .765,  .89}99.00\% & \cellcolor[rgb]{ .353,  .541,  .776}100.00\% & \cellcolor[rgb]{ .353,  .541,  .776}100.00\% & \cellcolor[rgb]{ .984,  .973,  .984}97.00\% & \cellcolor[rgb]{ .353,  .541,  .776}100.00\% \\
    \hline
    \multicolumn{1}{|c|}{3} & \cellcolor[rgb]{ .671,  .765,  .89}99.00\% & \cellcolor[rgb]{ .988,  .988,  1}98\% & \cellcolor[rgb]{ .984,  .965,  .976}96.50\% & \cellcolor[rgb]{ .353,  .541,  .776}100.00\% & \cellcolor[rgb]{ .988,  .988,  1}98.00\% & \cellcolor[rgb]{ .984,  .957,  .969}96.04\% & \cellcolor[rgb]{ .984,  .941,  .953}95.00\% & \cellcolor[rgb]{ .984,  .957,  .969}96.00\% & \cellcolor[rgb]{ .671,  .765,  .89}99.00\% & \cellcolor[rgb]{ .839,  .882,  .949}98.48\% & \cellcolor[rgb]{ .984,  .91,  .922}93.00\% & \cellcolor[rgb]{ .353,  .541,  .776}100.00\% \\
    \hline
    \multicolumn{1}{|c|}{4} & \cellcolor[rgb]{ .984,  .957,  .969}96.00\% & \cellcolor[rgb]{ .984,  .957,  .969}96.00\% & \cellcolor[rgb]{ .984,  .933,  .941}94.40\% & \cellcolor[rgb]{ .984,  .973,  .984}97.00\% & \cellcolor[rgb]{ .671,  .765,  .89}99.00\% & \cellcolor[rgb]{ .533,  .667,  .839}99.44\% & \cellcolor[rgb]{ .984,  .957,  .969}96.00\% & \cellcolor[rgb]{ .988,  .988,  1}98.00\% & \cellcolor[rgb]{ .984,  .973,  .984}97.00\% & \cellcolor[rgb]{ .984,  .969,  .98}96.72\% & \cellcolor[rgb]{ .984,  .878,  .89}91.00\% & \cellcolor[rgb]{ .988,  .988,  1}98.00\% \\
    \hline
    \multicolumn{1}{|c|}{5} & \cellcolor[rgb]{ .353,  .541,  .776}100.00\% & \cellcolor[rgb]{ .671,  .765,  .89}99.00\% & \cellcolor[rgb]{ .42,  .588,  .8}99.80\% & \cellcolor[rgb]{ .671,  .765,  .89}99.00\% & \cellcolor[rgb]{ .984,  .973,  .984}97.00\% & \cellcolor[rgb]{ .984,  .957,  .969}96.09\% & \cellcolor[rgb]{ .984,  .91,  .922}93.00\% & \cellcolor[rgb]{ .984,  .973,  .984}97.00\% & \cellcolor[rgb]{ .353,  .541,  .776}100.00\% & \cellcolor[rgb]{ .639,  .745,  .878}99.10\% & \cellcolor[rgb]{ .984,  .973,  .984}97.00\% & \cellcolor[rgb]{ .671,  .765,  .89}99.00\% \\
    \hline
    \multicolumn{1}{|c|}{6} & \cellcolor[rgb]{ .353,  .541,  .776}100.00\% & \cellcolor[rgb]{ .984,  .941,  .953}95.00\% & \cellcolor[rgb]{ .984,  .922,  .933}93.80\% & \cellcolor[rgb]{ .671,  .765,  .89}99.00\% & \cellcolor[rgb]{ .353,  .541,  .776}100.00\% & \cellcolor[rgb]{ .353,  .541,  .776}100.00\% & \cellcolor[rgb]{ .671,  .765,  .89}99.00\% & \cellcolor[rgb]{ .671,  .765,  .89}99.00\% & \cellcolor[rgb]{ .353,  .541,  .776}100.00\% & \cellcolor[rgb]{ .984,  .945,  .957}95.31\% & \cellcolor[rgb]{ .353,  .541,  .776}100.00\% & \cellcolor[rgb]{ .671,  .765,  .89}99.00\% \\
    \hline
    \multicolumn{1}{|c|}{7} & \cellcolor[rgb]{ .984,  .973,  .984}97.00\% & \cellcolor[rgb]{ .353,  .541,  .776}100.00\% & \cellcolor[rgb]{ .984,  .953,  .965}95.80\% & \cellcolor[rgb]{ .353,  .541,  .776}100.00\% & \cellcolor[rgb]{ .984,  .894,  .906}92.00\% & \cellcolor[rgb]{ .973,  .553,  .561}69.39\% & \cellcolor[rgb]{ .984,  .867,  .875}90.00\% & \cellcolor[rgb]{ .984,  .894,  .906}92.00\% & \cellcolor[rgb]{ .984,  .973,  .984}97.00\% & \cellcolor[rgb]{ .353,  .541,  .776}100.00\% & \cellcolor[rgb]{ .984,  .925,  .937}94.00\% & \cellcolor[rgb]{ .353,  .541,  .776}100.00\% \\
    \hline
    \multicolumn{1}{|c|}{8} & \cellcolor[rgb]{ .671,  .765,  .89}99.00\% & \cellcolor[rgb]{ .353,  .541,  .776}100.00\% & \cellcolor[rgb]{ .984,  .918,  .929}93.40\% & \cellcolor[rgb]{ .671,  .765,  .89}99.00\% & \cellcolor[rgb]{ .671,  .765,  .89}99.00\% & \cellcolor[rgb]{ .984,  .976,  .988}97.30\% & \cellcolor[rgb]{ .973,  .412,  .42}60.00\% & \cellcolor[rgb]{ .984,  .973,  .984}97.00\% & \cellcolor[rgb]{ .353,  .541,  .776}100.00\% & \cellcolor[rgb]{ .353,  .541,  .776}100.00\% & \cellcolor[rgb]{ .98,  .835,  .847}88.00\% & \cellcolor[rgb]{ .353,  .541,  .776}100.00\% \\
    \hline
    \multicolumn{1}{|c|}{9} & \cellcolor[rgb]{ .988,  .988,  1}98.00\% & \cellcolor[rgb]{ .984,  .878,  .89}91.00\% & \cellcolor[rgb]{ .353,  .541,  .776}100.00\% & \cellcolor[rgb]{ .988,  .988,  1}98.00\% & \cellcolor[rgb]{ .671,  .765,  .89}99.00\% & \cellcolor[rgb]{ .353,  .541,  .776}100.00\% & \cellcolor[rgb]{ .984,  .973,  .984}97.00\% & \cellcolor[rgb]{ .353,  .541,  .776}100.00\% & \cellcolor[rgb]{ .988,  .988,  1}98.00\% & \cellcolor[rgb]{ .984,  .894,  .906}91.89\% & \cellcolor[rgb]{ .984,  .91,  .922}93.00\% & \cellcolor[rgb]{ .988,  .988,  1}98.00\% \\
    \hline
    \multicolumn{1}{|c|}{10} & \cellcolor[rgb]{ .671,  .765,  .89}99.00\% & \cellcolor[rgb]{ .671,  .765,  .89}99.00\% & \cellcolor[rgb]{ .608,  .722,  .867}99.20\% & \cellcolor[rgb]{ .988,  .988,  1}98.00\% & \cellcolor[rgb]{ .988,  .988,  1}98.00\% & \cellcolor[rgb]{ .984,  .973,  .984}97.13\% & \cellcolor[rgb]{ .984,  .925,  .937}94.00\% & \cellcolor[rgb]{ .988,  .988,  1}98.00\% & \cellcolor[rgb]{ .671,  .765,  .89}99.00\% & \cellcolor[rgb]{ .51,  .651,  .831}99.51\% & \cellcolor[rgb]{ .984,  .925,  .937}94.00\% & \cellcolor[rgb]{ .671,  .765,  .89}99.00\% \\
    \hline
    \multicolumn{1}{|c|}{11} & \cellcolor[rgb]{ .353,  .541,  .776}100.00\% & \cellcolor[rgb]{ .988,  .988,  1}98.00\% & \cellcolor[rgb]{ .984,  .984,  .996}97.90\% & \cellcolor[rgb]{ .353,  .541,  .776}100.00\% & \cellcolor[rgb]{ .988,  .988,  1}98.00\% & \cellcolor[rgb]{ .831,  .878,  .945}98.50\% & \cellcolor[rgb]{ .984,  .894,  .906}92.00\% & \cellcolor[rgb]{ .671,  .765,  .89}99.00\% & \cellcolor[rgb]{ .353,  .541,  .776}100.00\% & \cellcolor[rgb]{ .671,  .765,  .89}99.00\% & \cellcolor[rgb]{ .671,  .765,  .89}99.00\% & \cellcolor[rgb]{ .353,  .541,  .776}100.00\% \\
    \hline
    \multicolumn{1}{|c|}{12} & \cellcolor[rgb]{ .671,  .765,  .89}99.00\% & \cellcolor[rgb]{ .984,  .941,  .953}95.00\% & \cellcolor[rgb]{ .831,  .878,  .945}98.50\% & \cellcolor[rgb]{ .671,  .765,  .89}99.00\% & \cellcolor[rgb]{ .353,  .541,  .776}100.00\% & \cellcolor[rgb]{ .518,  .659,  .835}99.49\% & \cellcolor[rgb]{ .984,  .894,  .906}92.00\% & \cellcolor[rgb]{ .671,  .765,  .89}99.00\% & \cellcolor[rgb]{ .353,  .541,  .776}100.00\% & \cellcolor[rgb]{ .984,  .949,  .961}95.61\% & \cellcolor[rgb]{ .984,  .957,  .969}96.00\% & \cellcolor[rgb]{ .353,  .541,  .776}100.00\% \\
    \hline
    \multicolumn{1}{|c|}{13} & \cellcolor[rgb]{ .353,  .541,  .776}100.00\% & \cellcolor[rgb]{ .671,  .765,  .89}99.00\% & \cellcolor[rgb]{ .984,  .965,  .976}96.70\% & \cellcolor[rgb]{ .671,  .765,  .89}99.00\% & \cellcolor[rgb]{ .988,  .988,  1}98.00\% & \cellcolor[rgb]{ .984,  .973,  .984}97.17\% & \cellcolor[rgb]{ .984,  .973,  .984}97.00\% & \cellcolor[rgb]{ .988,  .988,  1}98.00\% & \cellcolor[rgb]{ .353,  .541,  .776}100.00\% & \cellcolor[rgb]{ .506,  .651,  .831}99.52\% & \cellcolor[rgb]{ .984,  .851,  .859}89.00\% & \cellcolor[rgb]{ .353,  .541,  .776}100.00\% \\
    \hline
    \multicolumn{1}{|c|}{14} & \cellcolor[rgb]{ .988,  .988,  1}98.00\% & \cellcolor[rgb]{ .984,  .973,  .984}97.00\% & \cellcolor[rgb]{ .576,  .698,  .855}99.30\% & \cellcolor[rgb]{ .984,  .925,  .937}94.00\% & \cellcolor[rgb]{ .353,  .541,  .776}100.00\% & \cellcolor[rgb]{ .824,  .875,  .945}98.52\% & \cellcolor[rgb]{ .984,  .957,  .969}96.00\% & \cellcolor[rgb]{ .353,  .541,  .776}100.00\% & \cellcolor[rgb]{ .988,  .988,  1}98.00\% & \cellcolor[rgb]{ .984,  .98,  .992}97.56\% & \cellcolor[rgb]{ .984,  .894,  .906}92.00\% & \cellcolor[rgb]{ .984,  .941,  .953}95.00\% \\
    \hline
    \multicolumn{1}{|c|}{15} & \cellcolor[rgb]{ .353,  .541,  .776}100.00\% & \cellcolor[rgb]{ .988,  .988,  1}98.00\% & \cellcolor[rgb]{ .984,  .957,  .969}96.10\% & \cellcolor[rgb]{ .671,  .765,  .89}99.00\% & \cellcolor[rgb]{ .988,  .988,  1}98.00\% & \cellcolor[rgb]{ .851,  .89,  .953}98.44\% & \cellcolor[rgb]{ .984,  .878,  .89}91.00\% & \cellcolor[rgb]{ .988,  .988,  1}98.00\% & \cellcolor[rgb]{ .353,  .541,  .776}100.00\% & \cellcolor[rgb]{ .851,  .89,  .953}98.44\% & \cellcolor[rgb]{ .988,  .988,  1}98.00\% & \cellcolor[rgb]{ .671,  .765,  .89}99.00\% \\
    \hline
    \multicolumn{1}{|c|}{16} & \cellcolor[rgb]{ .988,  .988,  1}98.00\% & \cellcolor[rgb]{ .984,  .91,  .922}93.00\% & \cellcolor[rgb]{ .894,  .922,  .969}98.30\% & \cellcolor[rgb]{ .984,  .941,  .953}95.00\% & \cellcolor[rgb]{ .671,  .765,  .89}99.00\% & \cellcolor[rgb]{ .353,  .541,  .776}100.00\% & \cellcolor[rgb]{ .984,  .925,  .937}94.00\% & \cellcolor[rgb]{ .353,  .541,  .776}100.00\% & \cellcolor[rgb]{ .988,  .988,  1}98.00\% & \cellcolor[rgb]{ .984,  .922,  .933}93.75\% & \cellcolor[rgb]{ .984,  .973,  .984}97.00\% & \cellcolor[rgb]{ .984,  .941,  .953}95.00\% \\
    \hline
    \multicolumn{1}{|c|}{17} & \cellcolor[rgb]{ .671,  .765,  .89}99.00\% & \cellcolor[rgb]{ .671,  .765,  .89}99.00\% & \cellcolor[rgb]{ .984,  .906,  .918}92.80\% & \cellcolor[rgb]{ .353,  .541,  .776}100.00\% & \cellcolor[rgb]{ .984,  .973,  .984}97.00\% & \cellcolor[rgb]{ .984,  .929,  .941}94.20\% & \cellcolor[rgb]{ .984,  .957,  .969}96.00\% & \cellcolor[rgb]{ .984,  .957,  .969}96.00\% & \cellcolor[rgb]{ .671,  .765,  .89}99.00\% & \cellcolor[rgb]{ .518,  .659,  .835}99.49\% & \cellcolor[rgb]{ .984,  .925,  .937}94.00\% & \cellcolor[rgb]{ .353,  .541,  .776}100.00\% \\
    \hline
    \multicolumn{1}{|c|}{18} & \cellcolor[rgb]{ .671,  .765,  .89}99.00\% & \cellcolor[rgb]{ .988,  .988,  1}98.00\% & \cellcolor[rgb]{ .984,  .925,  .937}93.90\% & \cellcolor[rgb]{ .671,  .765,  .89}99.00\% & \cellcolor[rgb]{ .988,  .988,  1}98.00\% & \cellcolor[rgb]{ .667,  .761,  .886}99.02\% & \cellcolor[rgb]{ .988,  .988,  1}98.00\% & \cellcolor[rgb]{ .671,  .765,  .89}99.00\% & \cellcolor[rgb]{ .353,  .541,  .776}100.00\% & \cellcolor[rgb]{ .973,  .976,  .996}98.06\% & \cellcolor[rgb]{ .984,  .851,  .859}89.00\% & \cellcolor[rgb]{ .353,  .541,  .776}100.00\% \\
    \hline
    \multicolumn{1}{|c|}{19} & \cellcolor[rgb]{ .671,  .765,  .89}99.00\% & \cellcolor[rgb]{ .984,  .973,  .984}97.00\% & \cellcolor[rgb]{ .984,  .898,  .91}92.30\% & \cellcolor[rgb]{ .984,  .973,  .984}97.00\% & \cellcolor[rgb]{ .353,  .541,  .776}100.00\% & \cellcolor[rgb]{ .533,  .667,  .839}99.44\% & \cellcolor[rgb]{ .984,  .925,  .937}94.00\% & \cellcolor[rgb]{ .671,  .765,  .89}99.00\% & \cellcolor[rgb]{ .671,  .765,  .89}99.00\% & \cellcolor[rgb]{ .984,  .984,  .996}97.81\% & \cellcolor[rgb]{ .984,  .878,  .89}91.00\% & \cellcolor[rgb]{ .988,  .988,  1}98.00\% \\
    \hline
    \multicolumn{1}{|c|}{20} & \cellcolor[rgb]{ .984,  .973,  .984}97.00\% & \cellcolor[rgb]{ .984,  .894,  .906}92.00\% & \cellcolor[rgb]{ .984,  .933,  .945}94.60\% & \cellcolor[rgb]{ .984,  .957,  .969}96.00\% & \cellcolor[rgb]{ .671,  .765,  .89}99.00\% & \cellcolor[rgb]{ .506,  .651,  .831}99.52\% & \cellcolor[rgb]{ .984,  .941,  .953}95.00\% & \cellcolor[rgb]{ .671,  .765,  .89}99.00\% & \cellcolor[rgb]{ .984,  .973,  .984}97.00\% & \cellcolor[rgb]{ .984,  .91,  .918}92.83\% & \cellcolor[rgb]{ .984,  .878,  .89}91.00\% & \cellcolor[rgb]{ .984,  .957,  .969}96.00\% \\
    \hline
    \multicolumn{1}{|c|}{21} & \cellcolor[rgb]{ .671,  .765,  .89}99.00\% & \cellcolor[rgb]{ .988,  .988,  1}98.00\% & \cellcolor[rgb]{ .984,  .875,  .886}90.70\% & \cellcolor[rgb]{ .353,  .541,  .776}100.00\% & \cellcolor[rgb]{ .671,  .765,  .89}99.00\% & \cellcolor[rgb]{ .984,  .984,  .996}97.87\% & \cellcolor[rgb]{ .984,  .957,  .969}96.00\% & \cellcolor[rgb]{ .988,  .988,  1}98.00\% & \cellcolor[rgb]{ .671,  .765,  .89}99.00\% & \cellcolor[rgb]{ .863,  .902,  .957}98.40\% & \cellcolor[rgb]{ .984,  .894,  .906}92.00\% & \cellcolor[rgb]{ .353,  .541,  .776}100.00\% \\
    \hline
    \multicolumn{1}{|c|}{22} & \cellcolor[rgb]{ .988,  .988,  1}98.00\% & \cellcolor[rgb]{ .984,  .973,  .984}97.00\% & \cellcolor[rgb]{ .984,  .871,  .882}90.30\% & \cellcolor[rgb]{ .984,  .973,  .984}97.00\% & \cellcolor[rgb]{ .671,  .765,  .89}99.00\% & \cellcolor[rgb]{ .984,  .937,  .949}94.87\% & \cellcolor[rgb]{ .988,  .988,  1}98.00\% & \cellcolor[rgb]{ .671,  .765,  .89}99.00\% & \cellcolor[rgb]{ .988,  .988,  1}98.00\% & \cellcolor[rgb]{ .984,  .984,  .996}97.88\% & \cellcolor[rgb]{ .98,  .835,  .847}88.00\% & \cellcolor[rgb]{ .984,  .973,  .984}97.00\% \\
    \hline
    \multicolumn{1}{|c|}{23} & \cellcolor[rgb]{ .984,  .973,  .984}97.00\% & \cellcolor[rgb]{ .984,  .941,  .953}95.00\% & \cellcolor[rgb]{ .984,  .847,  .859}88.90\% & \cellcolor[rgb]{ .984,  .973,  .984}97.00\% & \cellcolor[rgb]{ .988,  .988,  1}98.00\% & \cellcolor[rgb]{ .663,  .761,  .886}99.03\% & \cellcolor[rgb]{ .671,  .765,  .89}99.00\% & \cellcolor[rgb]{ .984,  .973,  .984}97.00\% & \cellcolor[rgb]{ .984,  .973,  .984}97.00\% & \cellcolor[rgb]{ .984,  .953,  .965}95.79\% & \cellcolor[rgb]{ .984,  .957,  .969}96.00\% & \cellcolor[rgb]{ .984,  .973,  .984}97.00\% \\
    \hline
    \multicolumn{1}{|c|}{24} & \cellcolor[rgb]{ .988,  .988,  1}98.00\% & \cellcolor[rgb]{ .988,  .988,  1}98.00\% & \cellcolor[rgb]{ .984,  .855,  .867}89.40\% & \cellcolor[rgb]{ .671,  .765,  .89}99.00\% & \cellcolor[rgb]{ .671,  .765,  .89}99.00\% & \cellcolor[rgb]{ .678,  .773,  .894}98.98\% & \cellcolor[rgb]{ .984,  .894,  .906}92.00\% & \cellcolor[rgb]{ .671,  .765,  .89}99.00\% & \cellcolor[rgb]{ .671,  .765,  .89}99.00\% & \cellcolor[rgb]{ .918,  .937,  .976}98.23\% & \cellcolor[rgb]{ .353,  .541,  .776}100.00\% & \cellcolor[rgb]{ .671,  .765,  .89}99.00\% \\
    \hline
    \multicolumn{1}{|c|}{25} & \cellcolor[rgb]{ .984,  .973,  .984}97.00\% & \cellcolor[rgb]{ .984,  .957,  .969}96.00\% & \cellcolor[rgb]{ .984,  .886,  .898}91.40\% & \cellcolor[rgb]{ .984,  .973,  .984}97.00\% & \cellcolor[rgb]{ .988,  .988,  1}98.00\% & \cellcolor[rgb]{ .855,  .898,  .957}98.42\% & \cellcolor[rgb]{ .984,  .91,  .922}93.00\% & \cellcolor[rgb]{ .988,  .988,  1}98.00\% & \cellcolor[rgb]{ .988,  .988,  1}98.00\% & \cellcolor[rgb]{ .984,  .961,  .973}96.39\% & \cellcolor[rgb]{ .984,  .878,  .89}91.00\% & \cellcolor[rgb]{ .988,  .988,  1}98.00\% \\
    \hline
    \multicolumn{1}{|c|}{26} & \cellcolor[rgb]{ .671,  .765,  .89}99.00\% & \cellcolor[rgb]{ .353,  .541,  .776}100.00\% & \cellcolor[rgb]{ .984,  .902,  .914}92.50\% & \cellcolor[rgb]{ .353,  .541,  .776}100.00\% & \cellcolor[rgb]{ .671,  .765,  .89}99.00\% & \cellcolor[rgb]{ .984,  .973,  .984}97.22\% & \cellcolor[rgb]{ .984,  .973,  .984}97.00\% & \cellcolor[rgb]{ .671,  .765,  .89}99.00\% & \cellcolor[rgb]{ .671,  .765,  .89}99.00\% & \cellcolor[rgb]{ .353,  .541,  .776}100.00\% & \cellcolor[rgb]{ .984,  .973,  .984}97.00\% & \cellcolor[rgb]{ .353,  .541,  .776}100.00\% \\
    \hline
    \multicolumn{1}{|c|}{27} & \cellcolor[rgb]{ .988,  .988,  1}98.00\% & \cellcolor[rgb]{ .988,  .988,  1}98.00\% & \cellcolor[rgb]{ .984,  .871,  .882}90.50\% & \cellcolor[rgb]{ .984,  .973,  .984}97.00\% & \cellcolor[rgb]{ .671,  .765,  .89}99.00\% & \cellcolor[rgb]{ .984,  .922,  .933}93.85\% & \cellcolor[rgb]{ .984,  .973,  .984}97.00\% & \cellcolor[rgb]{ .671,  .765,  .89}99.00\% & \cellcolor[rgb]{ .988,  .988,  1}98.00\% & \cellcolor[rgb]{ .867,  .902,  .957}98.39\% & \cellcolor[rgb]{ .984,  .91,  .922}93.00\% & \cellcolor[rgb]{ .984,  .973,  .984}97.00\% \\
    \hline
    \multicolumn{1}{|c|}{28} & \cellcolor[rgb]{ .988,  .988,  1}98.00\% & \cellcolor[rgb]{ .984,  .973,  .984}97.00\% & \cellcolor[rgb]{ .98,  .839,  .851}88.30\% & \cellcolor[rgb]{ .984,  .973,  .984}97.00\% & \cellcolor[rgb]{ .353,  .541,  .776}100.00\% & \cellcolor[rgb]{ .353,  .541,  .776}100.00\% & \cellcolor[rgb]{ .984,  .941,  .953}95.00\% & \cellcolor[rgb]{ .984,  .941,  .953}95.00\% & \cellcolor[rgb]{ .988,  .988,  1}98.00\% & \cellcolor[rgb]{ .984,  .98,  .992}97.54\% & \cellcolor[rgb]{ .984,  .878,  .89}91.00\% & \cellcolor[rgb]{ .984,  .973,  .984}97.00\% \\
    \hline
    \multicolumn{1}{|c|}{29} & \cellcolor[rgb]{ .988,  .988,  1}98.00\% & \cellcolor[rgb]{ .984,  .957,  .969}96.00\% & \cellcolor[rgb]{ .984,  .871,  .882}90.50\% & \cellcolor[rgb]{ .984,  .957,  .969}96.00\% & \cellcolor[rgb]{ .671,  .765,  .89}99.00\% & \cellcolor[rgb]{ .686,  .776,  .894}98.96\% & \cellcolor[rgb]{ .988,  .988,  1}98.00\% & \cellcolor[rgb]{ .984,  .973,  .984}97.00\% & \cellcolor[rgb]{ .988,  .988,  1}98.00\% & \cellcolor[rgb]{ .984,  .969,  .98}96.95\% & \cellcolor[rgb]{ .984,  .973,  .984}97.00\% & \cellcolor[rgb]{ .984,  .973,  .984}97.00\% \\
    \hline
    \multicolumn{1}{|c|}{30} & \cellcolor[rgb]{ .671,  .765,  .89}99.00\% & \cellcolor[rgb]{ .988,  .988,  1}98.00\% & \cellcolor[rgb]{ .984,  .898,  .91}92.30\% & \cellcolor[rgb]{ .353,  .541,  .776}100.00\% & \cellcolor[rgb]{ .988,  .988,  1}98.00\% & \cellcolor[rgb]{ .89,  .918,  .965}98.32\% & \cellcolor[rgb]{ .984,  .851,  .859}89.00\% & \cellcolor[rgb]{ .984,  .925,  .937}94.00\% & \cellcolor[rgb]{ .671,  .765,  .89}99.00\% & \cellcolor[rgb]{ .89,  .918,  .965}98.32\% & \cellcolor[rgb]{ .353,  .541,  .776}100.00\% & \cellcolor[rgb]{ .353,  .541,  .776}100.00\% \\
    \hline
    \multicolumn{1}{|c|}{31} & \cellcolor[rgb]{ .671,  .765,  .89}99.00\% & \cellcolor[rgb]{ .671,  .765,  .89}99.00\% & \cellcolor[rgb]{ .984,  .886,  .898}91.50\% & \cellcolor[rgb]{ .988,  .988,  1}98.00\% & \cellcolor[rgb]{ .988,  .988,  1}98.00\% & \cellcolor[rgb]{ .957,  .965,  .988}98.11\% & \cellcolor[rgb]{ .984,  .867,  .875}90.00\% & \cellcolor[rgb]{ .988,  .988,  1}98.00\% & \cellcolor[rgb]{ .671,  .765,  .89}99.00\% & \cellcolor[rgb]{ .506,  .651,  .831}99.52\% & \cellcolor[rgb]{ .984,  .925,  .937}94.00\% & \cellcolor[rgb]{ .671,  .765,  .89}99.00\% \\
    \hline
    \multicolumn{1}{|c|}{32} & \cellcolor[rgb]{ .984,  .973,  .984}97.00\% & \cellcolor[rgb]{ .984,  .957,  .969}96.00\% & \cellcolor[rgb]{ .98,  .839,  .851}88.30\% & \cellcolor[rgb]{ .984,  .973,  .984}97.00\% & \cellcolor[rgb]{ .671,  .765,  .89}99.00\% & \cellcolor[rgb]{ .984,  .957,  .969}96.19\% & \cellcolor[rgb]{ .671,  .765,  .89}99.00\% & \cellcolor[rgb]{ .988,  .988,  1}98.00\% & \cellcolor[rgb]{ .984,  .973,  .984}97.00\% & \cellcolor[rgb]{ .984,  .965,  .976}96.65\% & \cellcolor[rgb]{ .98,  .835,  .847}88.00\% & \cellcolor[rgb]{ .984,  .973,  .984}97.00\% \\
    \hline
    \multicolumn{1}{|c|}{33} & \cellcolor[rgb]{ .984,  .973,  .984}97.00\% & \cellcolor[rgb]{ .984,  .973,  .984}97.00\% & \cellcolor[rgb]{ .98,  .808,  .82}86.20\% & \cellcolor[rgb]{ .984,  .957,  .969}96.00\% & \cellcolor[rgb]{ .353,  .541,  .776}100.00\% & \cellcolor[rgb]{ .984,  .976,  .988}97.38\% & \cellcolor[rgb]{ .984,  .973,  .984}97.00\% & \cellcolor[rgb]{ .353,  .541,  .776}100.00\% & \cellcolor[rgb]{ .984,  .973,  .984}97.00\% & \cellcolor[rgb]{ .984,  .976,  .988}97.38\% & \cellcolor[rgb]{ .984,  .91,  .922}93.00\% & \cellcolor[rgb]{ .984,  .957,  .969}96.00\% \\
    \hline
    \multicolumn{1}{|c|}{34} & \cellcolor[rgb]{ .988,  .988,  1}98.00\% & \cellcolor[rgb]{ .988,  .988,  1}98.00\% & \cellcolor[rgb]{ .984,  .851,  .859}89.00\% & \cellcolor[rgb]{ .984,  .973,  .984}97.00\% & \cellcolor[rgb]{ .984,  .973,  .984}97.00\% & \cellcolor[rgb]{ .984,  .886,  .898}91.37\% & \cellcolor[rgb]{ .988,  .988,  1}98.00\% & \cellcolor[rgb]{ .984,  .973,  .984}97.00\% & \cellcolor[rgb]{ .988,  .988,  1}98.00\% & \cellcolor[rgb]{ .875,  .91,  .961}98.36\% & \cellcolor[rgb]{ .984,  .925,  .937}94.00\% & \cellcolor[rgb]{ .988,  .988,  1}98.00\% \\
    \hline
    \multicolumn{1}{|c|}{35} & \cellcolor[rgb]{ .984,  .941,  .953}95.00\% & \cellcolor[rgb]{ .984,  .925,  .937}94.00\% & \cellcolor[rgb]{ .98,  .82,  .831}87.00\% & \cellcolor[rgb]{ .984,  .925,  .937}94.00\% & \cellcolor[rgb]{ .353,  .541,  .776}100.00\% & \cellcolor[rgb]{ .8,  .855,  .933}98.60\% & \cellcolor[rgb]{ .984,  .957,  .969}96.00\% & \cellcolor[rgb]{ .353,  .541,  .776}100.00\% & \cellcolor[rgb]{ .984,  .957,  .969}96.00\% & \cellcolor[rgb]{ .984,  .933,  .945}94.62\% & \cellcolor[rgb]{ .671,  .765,  .89}99.00\% & \cellcolor[rgb]{ .984,  .941,  .953}95.00\% \\
    \hline
    \multicolumn{1}{|c|}{36} & \cellcolor[rgb]{ .984,  .973,  .984}97.00\% & \cellcolor[rgb]{ .984,  .973,  .984}97.00\% & \cellcolor[rgb]{ .984,  .875,  .886}90.60\% & \cellcolor[rgb]{ .984,  .973,  .984}97.00\% & \cellcolor[rgb]{ .988,  .988,  1}98.00\% & \cellcolor[rgb]{ .984,  .957,  .965}95.94\% & \cellcolor[rgb]{ .671,  .765,  .89}99.00\% & \cellcolor[rgb]{ .988,  .988,  1}98.00\% & \cellcolor[rgb]{ .984,  .973,  .984}97.00\% & \cellcolor[rgb]{ .984,  .976,  .988}97.42\% & \cellcolor[rgb]{ .984,  .957,  .969}96.00\% & \cellcolor[rgb]{ .988,  .988,  1}98.00\% \\
    \hline
    \multicolumn{1}{|c|}{37} & \cellcolor[rgb]{ .988,  .988,  1}98.00\% & \cellcolor[rgb]{ .988,  .988,  1}98.00\% & \cellcolor[rgb]{ .98,  .812,  .82}86.40\% & \cellcolor[rgb]{ .984,  .941,  .953}95.00\% & \cellcolor[rgb]{ .671,  .765,  .89}99.00\% & \cellcolor[rgb]{ .988,  .988,  1}98.00\% & \cellcolor[rgb]{ .984,  .941,  .953}95.00\% & \cellcolor[rgb]{ .671,  .765,  .89}99.00\% & \cellcolor[rgb]{ .671,  .765,  .89}99.00\% & \cellcolor[rgb]{ .835,  .882,  .949}98.49\% & \cellcolor[rgb]{ .984,  .851,  .859}89.00\% & \cellcolor[rgb]{ .984,  .941,  .953}95.00\% \\
    \hline
    \multicolumn{1}{|c|}{38} & \cellcolor[rgb]{ .988,  .988,  1}98.00\% & \cellcolor[rgb]{ .988,  .988,  1}98.00\% & \cellcolor[rgb]{ .98,  .839,  .847}88.20\% & \cellcolor[rgb]{ .984,  .941,  .953}95.00\% & \cellcolor[rgb]{ .353,  .541,  .776}100.00\% & \cellcolor[rgb]{ .984,  .969,  .98}96.81\% & \cellcolor[rgb]{ .984,  .894,  .906}92.00\% & \cellcolor[rgb]{ .671,  .765,  .89}99.00\% & \cellcolor[rgb]{ .988,  .988,  1}98.00\% & \cellcolor[rgb]{ .871,  .906,  .961}98.38\% & \cellcolor[rgb]{ .984,  .894,  .906}92.00\% & \cellcolor[rgb]{ .984,  .941,  .953}95.00\% \\
    \hline
    Avg(1Defect) & \cellcolor[rgb]{ .788,  .851,  .933}98.63\% & \cellcolor[rgb]{ .984,  .976,  .988}97.29\% & \cellcolor[rgb]{ .984,  .961,  .973}96.44\% & \cellcolor[rgb]{ .671,  .765,  .89}99.00\% & \cellcolor[rgb]{ .984,  .984,  .996}97.88\% & \cellcolor[rgb]{ .984,  .937,  .949}94.72\% & \cellcolor[rgb]{ .984,  .871,  .882}90.38\% & \cellcolor[rgb]{ .984,  .976,  .988}97.25\% & \cellcolor[rgb]{ .71,  .792,  .902}98.88\% & \cellcolor[rgb]{ .984,  .984,  .996}97.95\% & \cellcolor[rgb]{ .984,  .922,  .933}93.78\% & \cellcolor[rgb]{ .569,  .694,  .855}99.33\% \\
    \hline
    Avg(2Defects) & \cellcolor[rgb]{ .722,  .8,  .906}98.85\% & \cellcolor[rgb]{ .984,  .969,  .98}96.92\% & \cellcolor[rgb]{ .984,  .949,  .957}95.43\% & \cellcolor[rgb]{ .984,  .984,  .996}97.92\% & \cellcolor[rgb]{ .773,  .835,  .925}98.69\% & \cellcolor[rgb]{ .988,  .988,  1}98.01\% & \cellcolor[rgb]{ .984,  .937,  .949}94.85\% & \cellcolor[rgb]{ .792,  .851,  .933}98.62\% & \cellcolor[rgb]{ .671,  .765,  .89}99.00\% & \cellcolor[rgb]{ .984,  .98,  .992}97.53\% & \cellcolor[rgb]{ .984,  .91,  .922}93.08\% & \cellcolor[rgb]{ .871,  .906,  .961}98.38\% \\
    \hline
    Avg(3Defects) & \cellcolor[rgb]{ .984,  .984,  .996}97.92\% & \cellcolor[rgb]{ .984,  .976,  .988}97.33\% & \cellcolor[rgb]{ .984,  .863,  .875}89.90\% & \cellcolor[rgb]{ .984,  .98,  .992}97.58\% & \cellcolor[rgb]{ .776,  .839,  .925}98.67\% & \cellcolor[rgb]{ .984,  .976,  .988}97.32\% & \cellcolor[rgb]{ .984,  .945,  .957}95.33\% & \cellcolor[rgb]{ .984,  .98,  .992}97.58\% & \cellcolor[rgb]{ .965,  .973,  .992}98.08\% & \cellcolor[rgb]{ .984,  .984,  .996}97.79\% & \cellcolor[rgb]{ .984,  .933,  .945}94.50\% & \cellcolor[rgb]{ .984,  .984,  .996}97.92\% \\
    \hline
    Avg(4Defects) & \cellcolor[rgb]{ .984,  .973,  .984}97.00\% & \cellcolor[rgb]{ .984,  .969,  .98}96.75\% & \cellcolor[rgb]{ .98,  .835,  .847}88.05\% & \cellcolor[rgb]{ .984,  .945,  .957}95.25\% & \cellcolor[rgb]{ .592,  .71,  .863}99.25\% & \cellcolor[rgb]{ .984,  .976,  .988}97.34\% & \cellcolor[rgb]{ .984,  .949,  .961}95.50\% & \cellcolor[rgb]{ .671,  .765,  .89}99.00\% & \cellcolor[rgb]{ .984,  .98,  .992}97.50\% & \cellcolor[rgb]{ .984,  .976,  .988}97.23\% & \cellcolor[rgb]{ .984,  .925,  .937}94.00\% & \cellcolor[rgb]{ .984,  .953,  .965}95.75\% \\
    \hline
    \multicolumn{1}{|c|}{Avg(All38)} & \cellcolor[rgb]{ .882,  .914,  .965}98.34\% & \cellcolor[rgb]{ .984,  .973,  .984}97.19\% & \cellcolor[rgb]{ .984,  .914,  .925}93.20\% & \cellcolor[rgb]{ .925,  .945,  .98}98.20\% & \cellcolor[rgb]{ .796,  .855,  .933}98.61\% & \cellcolor[rgb]{ .984,  .973,  .984}97.08\% & \cellcolor[rgb]{ .984,  .925,  .937}94.00\% & \cellcolor[rgb]{ .965,  .973,  .992}98.08\% & \cellcolor[rgb]{ .816,  .867,  .941}98.55\% & \cellcolor[rgb]{ .984,  .98,  .992}97.68\% & \cellcolor[rgb]{ .984,  .941,  .953}95.00\% & \cellcolor[rgb]{ .933,  .949,  .98}98.18\% \\
    \hline    
    \end{tabular}%
  \label{tab:classwise}%
\end{table*}%

WSCN achieves an average classification accuracy of 98.20\% on all 38 classes. For single, two mixed-type, three mixed-type, and four mixed-type defect patterns, the average classification accuracy of WSCN are 99.0\%, 97.9\%, 97.6\% and 95.3\%, respectively. Thus, with increasing mixed-type defects, the accuracy degrades gracefully. This is expected since three or four mixed-type defect patterns are difficult to classify, even for humans. For some of these images, the ground-truth (GT) label provided in the dataset itself appears to be confusing (refer Section \ref{sec:mispredictedCases}).

Table \ref{tab:classificationResultsSummary} summarizes the results on classification accuracy and also presents the Matthews correlation coefficient (MCC), the number of parameters, model size and number of computations (FLOPS) to allow comparison from different perspectives. We evaluate both small models such as LeNet, MobileNetV2, EfficientNetB0 and ResNet18 and large models such as AlexNet, ResNet50 and DenseNet121. This is to cover networks with high accuracy and networks with low model size.
 
The average classification accuracy of DCNet and ResNet50 are 93.20\% and 97.19\%, respectively. Thus, WSCN achieves superior results than DCNet and ResNet50. The classification accuracy of WSCN is marginally lower than that of DenseNet (98.34\%). 
WSCN has only 0.09M parameters, a model size of 0.51MB, and a computation count of 0.2M.  
On analyzing the MCC values, we note that WSCN (with contrastive loss) achieves higher MCC value than other networks, except for DenseNet121. 
 A well-known limitation of DenseNet121 is that during inference, its memory consumption increases quadratically with network depth. Hence, it has a huge memory footprint (sum of model size and memory consumed by intermediate activations), leading to out-of-memory error. 
WSCN with contrastive loss has 0.09M parameters, but WSCN without contrastive loss has 0.08M parameters. The increase in the number of parameters is due to addition of a dense layer for generating required shape in latent dimension while using the contrastive loss.
So considering the balance of model size and FLOPS, our proposed model WSCN can be regarded as the best model. In fact, the model size of WSCN is 200$\times$ lower than that of ResNet50.

Among the lighweight models, LeNet-5 has lower accuracy of 76.44\%, while AlexNet and MobileNetV2 both shows accuracy of 97.91\%. EfficientNetB0 has slightly lower accuracy of 96.87\%. ResNet18 gives slightly better results than ResNet50, which is possible due to the simplicity of wafermap dataset.

\begin{table}[htbp]
\centering
\caption{Comparison of classification accuracy, MCC, parameters, model size and FLOPS of different models }\label{tab:classificationResultsSummary}
\begin{tabu}{|l|c|c|c|c|c|}
\hline
Model             & Accuracy  &   MCC  & Params   & Model Size (MB) & FLOPS \\ \hline
DenseNet121 \cite{huang2017densely}    & 98.34\%    &  0.9869  & 8M     & 33.48     & 16.1M  \\ \hline
ResNet18            					  &97.96 \% & 0.9791 & 11.2M &  44.98 &   22.39M   \\ \hline
ResNet50 \cite{he2016deep}             & 97.19\%    &  0.9758  & 25.7M  & 103.34    & 51.3M \\ \hline
DCNet \cite{wang2020deformable}		   & 93.20\%    & -      & 	2.3M  &	9.09 	  &	1.4M	\\ \hline
 LeNet \cite{lecun1998gradient}         & 76.44\%    &  0.7581  &  5.9M  & 23.96     & 12.0M   \\ \hline
 AlexNet \cite{krizhevsky2012imagenet}  & 97.91\%    &  0.9785  &  7.6M  & 30.47     & 15.2M   \\ \hline
MobileNetV2 \cite{sandler2018mobilenetv2} &97.91\%  &  0.9679  &  3.6M  & 14.83     &  7.1M    \\ \hline
EfficientNetB0 \cite{tan2019efficientnet} & 96.87\% &  0.9620  &  5.4M  & 22.14     & 10.7M    \\ \hline
WSCN (with BCE loss) 					  & 93.29\% &  0.9622	 & 0.08M  & 0.51      & 0.2M \\ \hline
WSCN (with contrastive loss)              & 98.20\% & 0.9815   & 0.09M  & 0.51 & 0.2M \\ \hline
\end{tabu}
\end{table}

WSCN achieves a precision value of 98.08\% and a recall value of 98.18\%. As shown in Table \ref{tab:classwise}, these precision and recall values are much superior to that of DCNet, better than that of ResNet50, and only slightly lower than that of DenseNet121. The values above 98\% indicate that WSCN is highly effective. As we show in Section \ref{sec:mispredictedCases}, careful re-evaluation of GT labels on the misclassified data shows confusing GT labels. This has led to mispredictions which reflected in the recall/precision values.

\textbf{Comparison with DCNet \cite{wang2020deformable}:} DCNet has used stochastic gradient descent (SGD), which is slow in convergence. In fact, DCNet requires 4,000 epochs to reach 93.2\% accuracy. This indicates that the SGD optimizer is not optimal for WM classification. By contrast, we have used Adam optimizer with N-pair contrastive loss. This allows WSCN to converge in 150 epochs only, reducing the time to train. N-pair contrastive loss helps in better extraction of feature representations of the input image in the encoder. This also improves classification accuracy. As shown in Table \ref{tab:classificationResultsSummary}, on using a different loss such as the conventional BCE loss, the accuracy of WSCN reduces to 93.29\%.

\subsection{ROC-AUC curve} \label{sec:ROC}
 
\begin{figure}[htbp]  \centering
\includegraphics [scale=0.70] {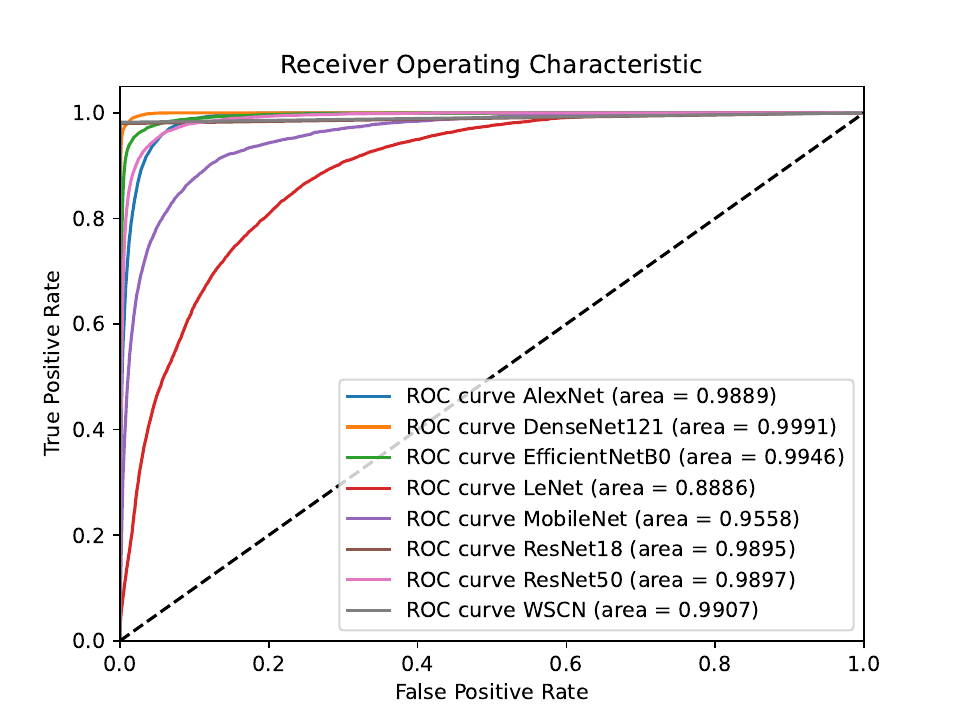}
\caption{ROC-AUC curve for different classifiers}
 \label{fig:ROC}
  \end{figure}

Figure \ref{fig:ROC} shows the ROC curve for performance evaluation of different classification models.   WSCN has an AUC of 0.9907, which is marginally below that of DenseNet121 (0.9991) and EfficientNetB0 (0.9946). All other classifiers have lower AUC value than WSCN.  

ROC-AUC curve depends on both the true positive rate and the false positive rate. However, the true negative and the false negative rate information is neglected in ROC-AUC curve. 
Thus, the ROC-AUC curve may provide a misleading impression that the model gives good performance even though there might be chances of misclassification. 
In Table \ref{tab:classificationResultsSummary}, we have presented the MCC metric, which gives a more complete picture of a classifier's performance than the ROC-AUC curve. 

\subsection{Analysis of mispredicted wafer-maps}\label{sec:mispredictedCases}
\begin{figure}[htbp]
        \centering
        \begin{subfigure}[b]{0.23\textwidth}
            \centering
            \includegraphics[width=0.9\textwidth]{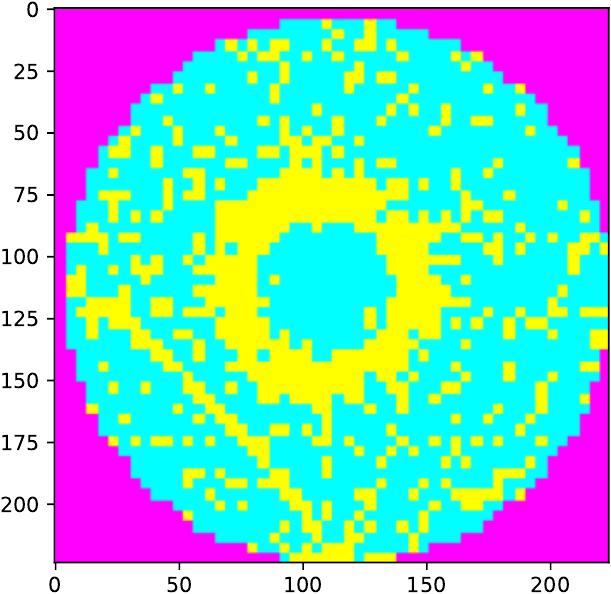}
            \caption[Network2]%
            {{\small GT:D+EL+S, Pred: D+S}}    
            \label{fig:Misclassified1}
        \end{subfigure}
        %\hfill
        \begin{subfigure}[b]{0.23\textwidth}  
            \centering 
            \includegraphics[width=0.9\textwidth]{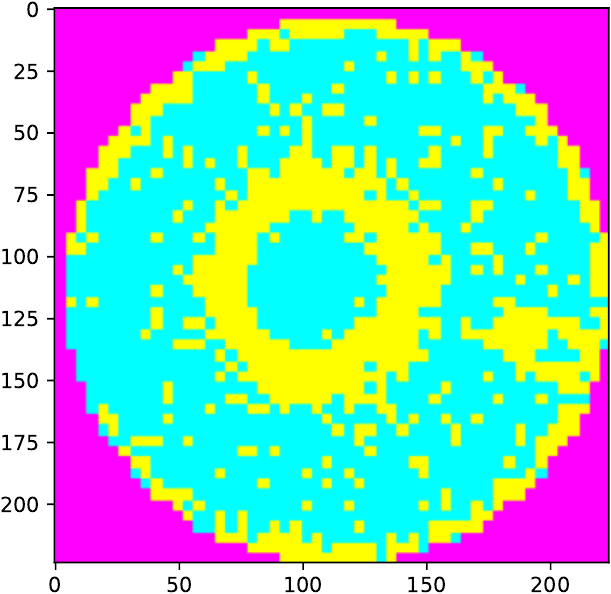}
            \caption[]%
            {{\small GT:D+EL+L, Pred:D+ER+L}}    
            \label{fig:Misclassified2}
        \end{subfigure}
        %\vskip\baselineskip
        \begin{subfigure}[b]{0.23\textwidth}   
            \centering 
            \includegraphics[width=0.9\textwidth]{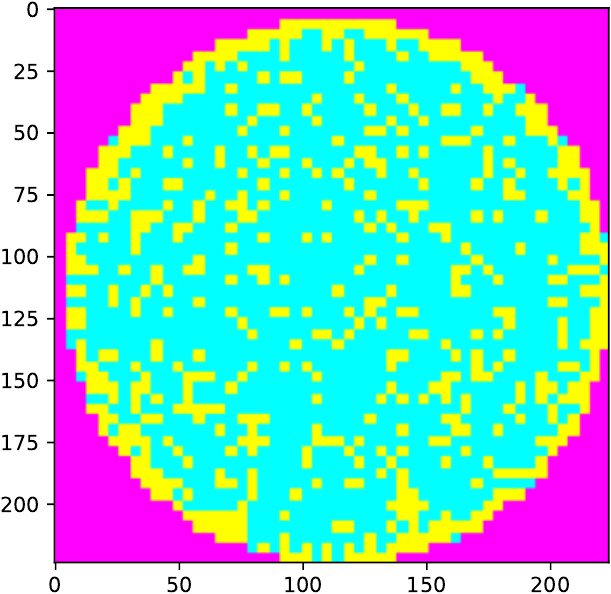}
            \caption[]%
            {{\small GT:ER+L, Pred: ER}}    
            \label{fig:Misclassified3}
        \end{subfigure}
        %\hfill
        \begin{subfigure}[b]{0.23\textwidth}   
            \centering 
            \includegraphics[width=0.9\textwidth]{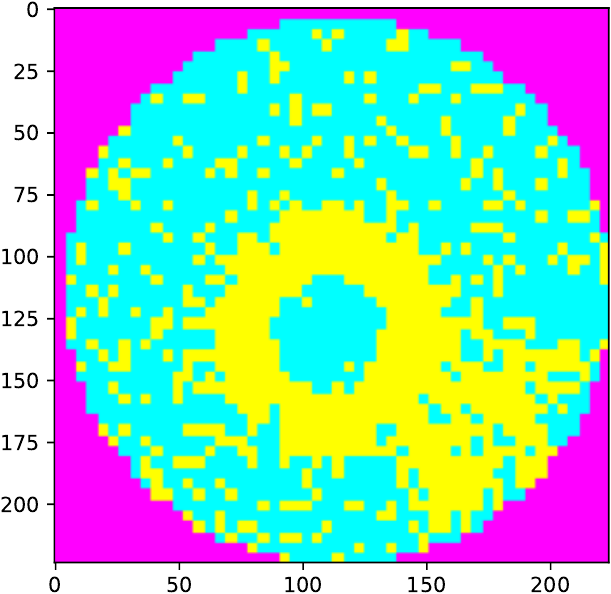}
            \caption[]%
            {{\small GT: D+L, Pred: D+EL+L}}    
            \label{fig:Misclassified4}
        \end{subfigure}
        \caption[]
        {\small Four sample failure cases (GT=ground-truth, Pred = prediction)} 
        \label{fig:Misclassified}
    \end{figure}

We further investigate the 138 WMs that our model misclassifies. Figure \ref{fig:Misclassified} shows four such sample WMs. 
In Figure \ref{fig:Misclassified1}, the GT annotation is Donut + EdgeLoc + Scratch. However, as evident from the figure and as predicted by our model, the GT should be Donut + Scratch only. Similarly, for Figures \ref{fig:Misclassified2} and \ref{fig:Misclassified3}, we find that the GT labels in the dataset are D + EL + L and ER + L, respectively, but our model predicts D + ER + L and ER, respectively. Our predictions appear to be more reasonable according to the visible defect pattern.
For a few other mispredicted WMs also, we observe that the GT labels appear to be confusing, whereas our model predicts correctly.

For most misclassified 138 images, it is difficult for the model to learn the defects from the GT labels as there is no uniformity across the defect images. Especially, given the high similarity between EL and ER, distinguishing between them is challenging, and the correct GT label itself is debatable. Similarly, every ``nearfull'' defect is also a random defect. Hence, for four out of 138 images, the GT is ``random'', whereas the model predicts it as ``nearfull'' defect. Further, any localized defect, regardless of its size, is considered as L. Hence, it is difficult to distinguish an L defect and a regular (fault-free) pattern or other defect patterns. This discrepancy can be addressed if we have a dataset containing well-annotated bounding boxes or masks validated by semiconductor manufacturing experts.

For some WMs, such as the one shown in  Figure \ref{fig:Misclassified4}, our model gives incorrect prediction. We believe that these WMs are difficult to predict even for humans.

\begin{table*}[htbp]
  \centering
  \caption{Defect-class, corresponding input image, ground truth and predicted mask generated by WSCN}\label{tab:segmentationVisualization}
  \begin{tabular}{ | c | C{3.75cm} | C{3.75cm} | C{3.75cm} | }
    \hline
    Defect Type & Input Image & Ground Truth & Generated Mask \\ \hline
    	Center
    &
    %\begin{minipage}[t]{5cm}
      \begin{minipage}{.2\textwidth}
      \includegraphics[width=\linewidth, height=30mm]{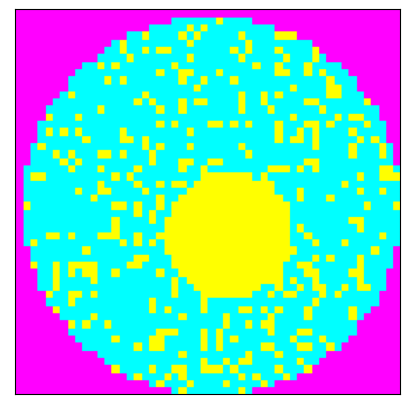}
    \end{minipage}
    %\end{minipage}
    & 
    %\begin{minipage}{5cm}
      \begin{minipage}{.2\textwidth}
      \includegraphics[width=\linewidth, height=30mm]{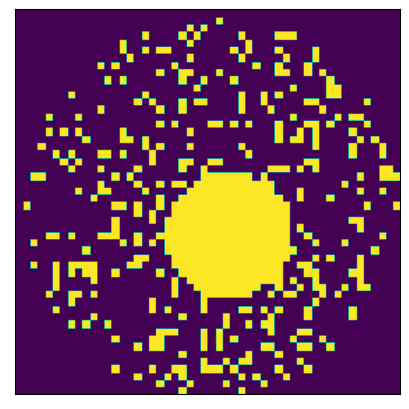}
    \end{minipage}
    %\end{minipage} 
    &
    %\begin{minipage}[t]{5cm}
      \begin{minipage}{.2\textwidth}
      \includegraphics[width=\linewidth, height=30mm]{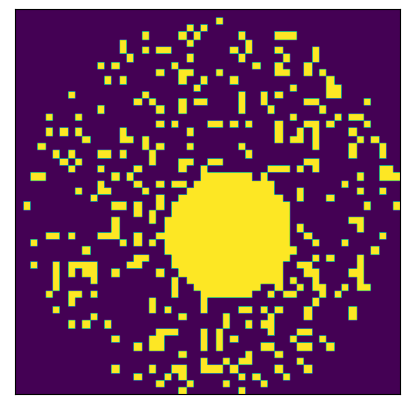}
    \end{minipage}  \\ \hline
		Donut
	&
		\begin{minipage}{.2\textwidth}
      \includegraphics[width=\linewidth, height=30mm]{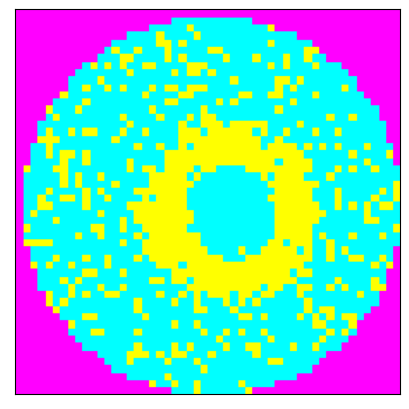}
    \end{minipage}
	&
		\begin{minipage}{.2\textwidth}
      \includegraphics[width=\linewidth, height=30mm]{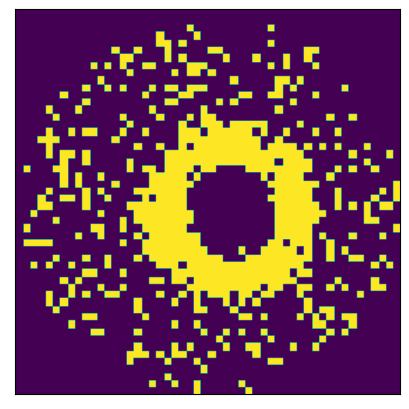}
    \end{minipage}
	&    
    	\begin{minipage}{.2\textwidth}
      \includegraphics[width=\linewidth, height=30mm]{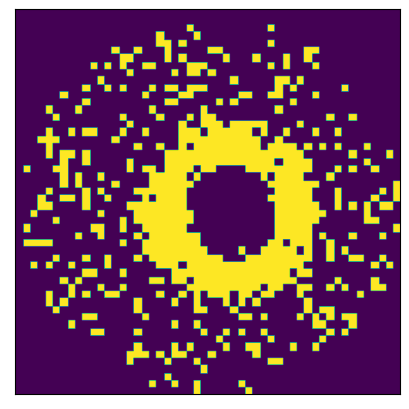}
    \end{minipage}  \\ \hline
		edgeLoc
	&
		\begin{minipage}{.2\textwidth}
      \includegraphics[width=\linewidth, height=30mm]{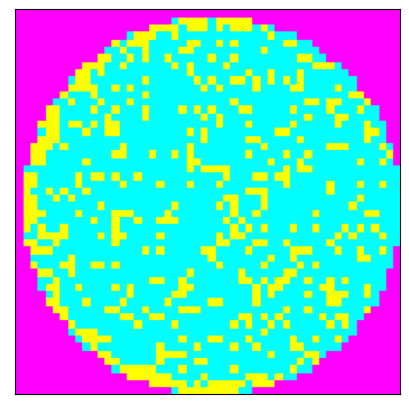}
    \end{minipage}
	&
		\begin{minipage}{.2\textwidth}
      \includegraphics[width=\linewidth, height=30mm]{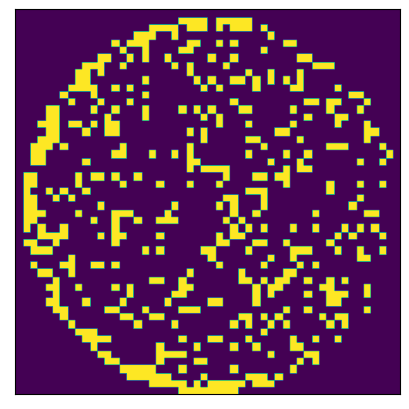}
    \end{minipage}
	&    
    	\begin{minipage}{.2\textwidth}
      \includegraphics[width=\linewidth, height=30mm]{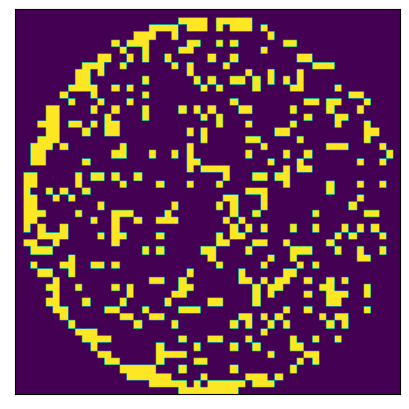}
    \end{minipage}     \\ \hline
		edgeRing
	&
		\begin{minipage}{.2\textwidth}
      \includegraphics[width=\linewidth, height=30mm]{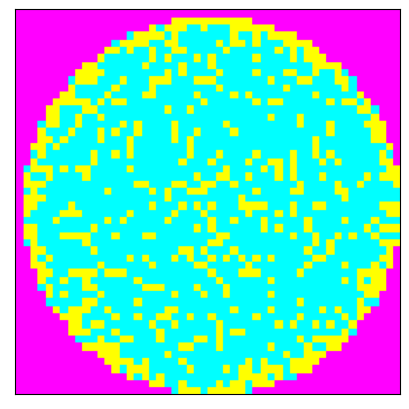}
    \end{minipage}
	&
		\begin{minipage}{.2\textwidth}
      \includegraphics[width=\linewidth, height=30mm]{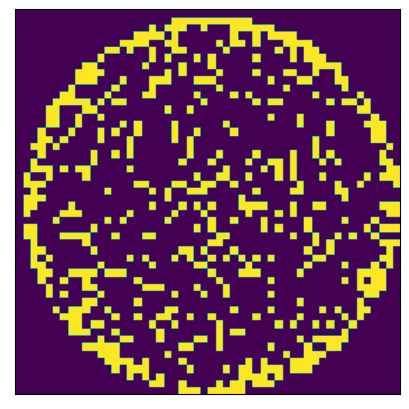}
    \end{minipage}
	&    
    	\begin{minipage}{.2\textwidth}
      \includegraphics[width=\linewidth, height=30mm]{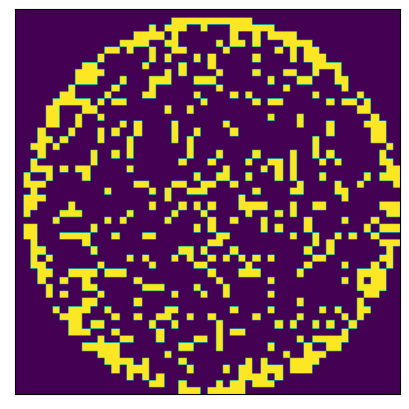}
    \end{minipage}   \\ \hline
    
    Loc
	&
		\begin{minipage}{.2\textwidth}
      \includegraphics[width=\linewidth, height=30mm]{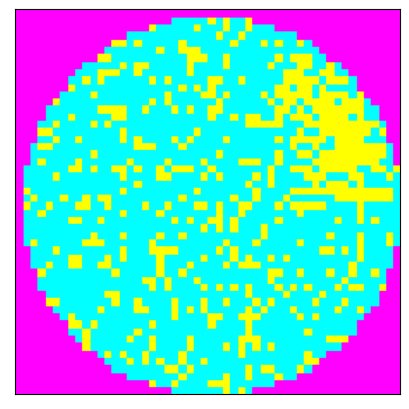}
    \end{minipage}
	&
		\begin{minipage}{.2\textwidth}
      \includegraphics[width=\linewidth, height=30mm]{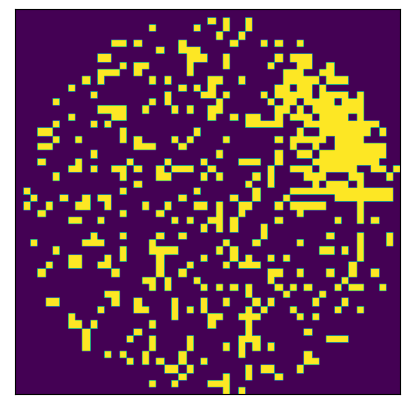}
    \end{minipage}
	&    
    	\begin{minipage}{.2\textwidth}
      \includegraphics[width=\linewidth, height=30mm]{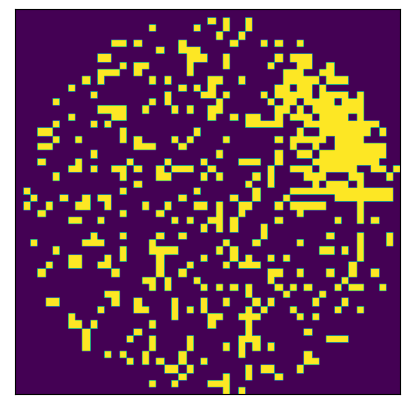}
    \end{minipage} \\ \hline
		Nearful
	&
		\begin{minipage}{.2\textwidth}
      \includegraphics[width=\linewidth, height=30mm]{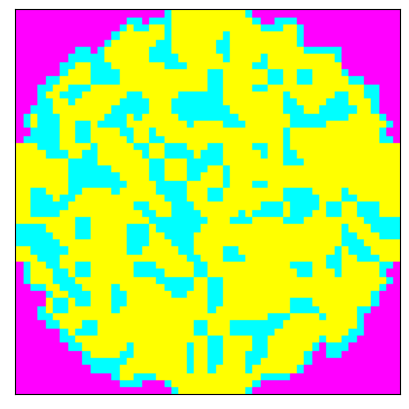}
    \end{minipage}
	&
		\begin{minipage}{.2\textwidth}
      \includegraphics[width=\linewidth, height=30mm]{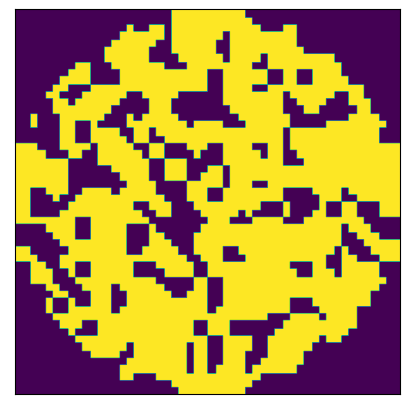}
    \end{minipage}
	&    
    	\begin{minipage}{.2\textwidth}
      \includegraphics[width=\linewidth, height=30mm]{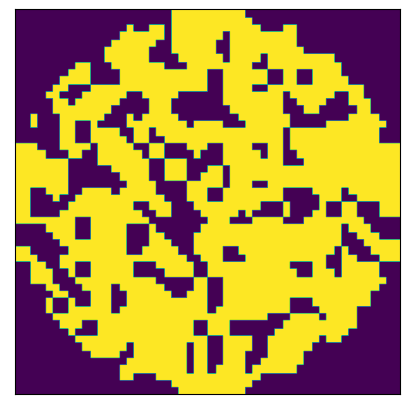}
    \end{minipage}   \\ \hline
		Scratch
	&
		\begin{minipage}{.2\textwidth}
      \includegraphics[width=\linewidth, height=30mm]{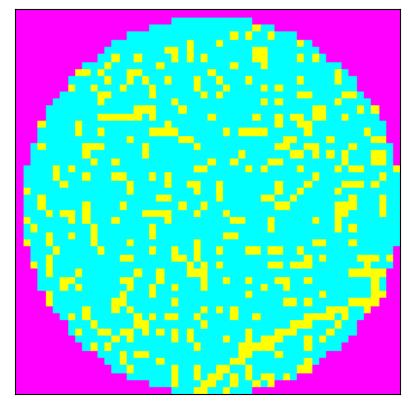}
    \end{minipage}
	&
		\begin{minipage}{.2\textwidth}
      \includegraphics[width=\linewidth, height=30mm]{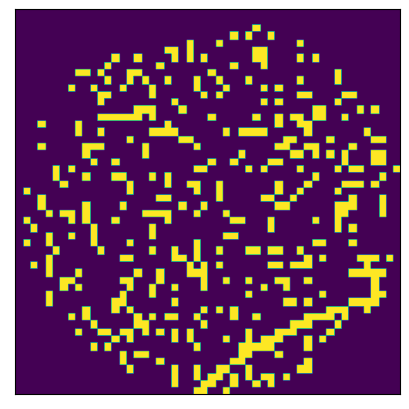}
    \end{minipage}
	&    
    	\begin{minipage}{.2\textwidth}
      \includegraphics[width=\linewidth, height=30mm]{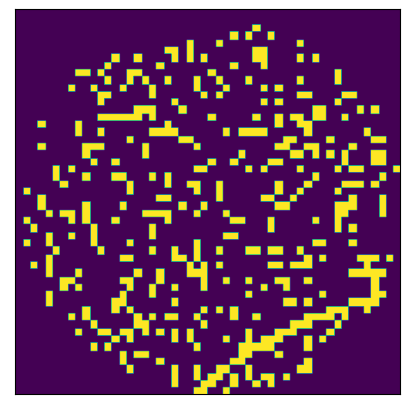}
    \end{minipage} \\ \hline
		Random
	&
		\begin{minipage}{.2\textwidth}
      \includegraphics[width=\linewidth, height=30mm]{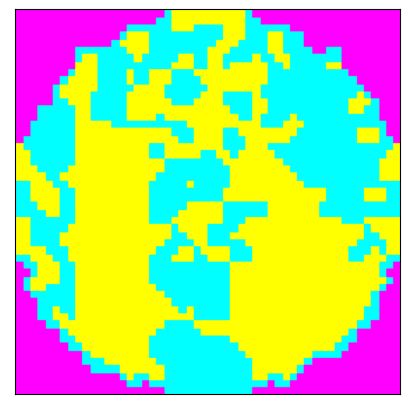}
    \end{minipage}
	&
		\begin{minipage}{.2\textwidth}
      \includegraphics[width=\linewidth, height=30mm]{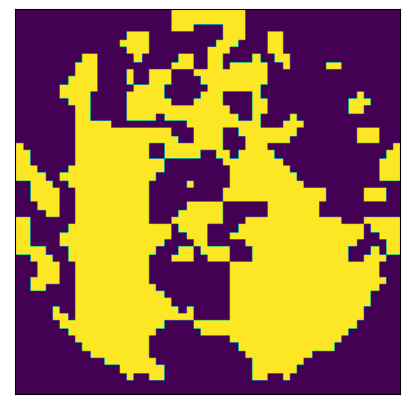}
    \end{minipage}
	&    
    	\begin{minipage}{.2\textwidth}
      \includegraphics[width=\linewidth, height=30mm]{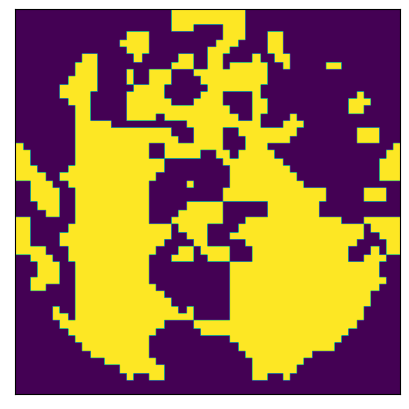}
    \end{minipage} \\ \hline
  \end{tabular}
\end{table*}

\subsection{Segmentation Results}\label{sec:segmentation}
We have used ``IOU'' and ``dice coefficient'' to measure the performance of our model. 
``Dice coefficient'' demonstrates the similarity between the ``ground truth'' and ``predicted mask''. As shown in Table \ref{tab:quantsegresults} in the ``WSCN(FP32)'' row, the value of dice-coefficient is 0.9999. This value indicates an extremely high  similarity between ground truth and predicted mask. WSCN achieves an IoU of 0.9999, which confirms its high segmentation capability.

\begin{table}[htbp]
\centering
\caption{Model size and segmentation results of open-source models, WSCN(FP32), i.e., without quantization and WSCN(INT8), i.e., with INT8 quantization}\label{tab:quantsegresults}
\begin{tabular}{|l|c|c|c|}\hline
                               & Model Size & Dice Coefficient & IOU \\ \hline
UNet \cite{ronneberger2015u}                           &  125.77MB   & 0.9998          & 0.9881 \\ \hline
DeepLabV3+ (ResNet50 backbone) \cite{chen2017deeplab} & 47.83 MB    & 0.9946          & 0.9924 \\ \hline 
WSCN(FP32)              &   0.51 MB   & 0.9999          & 0.9999      \\ \hline
WSCN(INT8)              &  0.17MB     & 0.9999            & 0.9999     \\ \hline 

\end{tabular}
\end{table}
 
\textbf{Comparative evaluation:} Since ours is the first model to show segmentation on the MixedWM38 dataset, we compare WSCN against two well-known open-source segmentation models, namely UNet and DeepLabV3+ (with ResNet50 backbone). Table \ref{tab:quantsegresults} shows the results. While these models achieve comparable results as WSCN, their model size is much larger. In fact, WSCN's model size is  250$\times$ lower than that of UNet. Given this, WSCN can be considered the best among the segmentation models. Since the dice coefficient and IoU values of WSCN are 0.9999, the ROC-AUC curve  remains close to 1. Hence, we omit it

\textbf{Visualization of WSCN results:} Table \ref{tab:segmentationVisualization} shows selected input images, corresponding ground-truth images, and the binary segmentation masks generated by WSCN. Evidently, the mask generated by WSCN is identical to the ground truth.

\subsection{Latency and quantization results}\label{sec:latency}

\textbf{Inference latency of WSCN:} We measure the latency of WSCN execution as follows. We first resize the input image from $52\times52$ to $224\times224$ size. We compute results of each image in batch of 1000 images from test set by sending corresponding image to WSCN, our  proposed model.  We record the time taken by each image for completing this process. We repeat this process ten times and take the average.
We observe a frame-rate of 25.11 frames-per-second on the Tesla P100 GPU. Thus, WSCN can work in real-time to perform defect classification with high-performance.

\textbf{Reducing model size using quantization:} The reference WSCN model uses single-precision (FP32) weights. We further reduce the model size by performing full integer quantization. The representative dataset is used to determine the min-max values of the inputs in full-integer quantization. When the converter quantizes the model, these are required to appropriately calculate the quantization nodes. As shown in Table \ref{tab:quantsegresults}, quantization reduces the model size of WSCN to merely 0.17MB, with negligible impact on the dice-coefficient and IoU values.

\subsection{Limitations of Proposed Methodology}\label{sec:limitation}
We note the following areas for further improvement of our approach:

1. \textbf{Performance on unseen classes:} 
We have used predetermined defect classes and combinations of multiple defect classes. Due to the dynamic nature of the manufacturing process, all the different types of defects cannot be decided in the initial stage. We have used a CNN-based dynamic defect feature extraction approach from wafermap images. Since WSCN is trained on predetermined classes, it may show lower performance on real-world defect classes that are not introduced to the model in the training phase. To achieve high accuracy on those defects, we would need to retrain or fine-tune the model. Also, we plan to train WSCN using unsupervised learning to further improve its classification performance. 

2. \textbf{Ability to distinguish \textit{similar-looking} unseen classes:} We have trained our feature extraction encoder using N-pair contrastive loss. It helps in finding similarity between embeddings from the same class in the training phase.  If our model comes across an unknown class which is similar to one of the known classes on which the model has been trained, our model would predict that unknown class to the similar  known class. For example, if there is an unknown class which is similar to ER, then our model would predict the unknown class as ER defect. This is due to the fact that N-pair contrastive loss in the encoder maps the embeddings of similar patterns of defects close to each other in the latent representation.

3. \textbf{Improving training efficiency:} Currently, we have trained WSCN model from scratch. To increase the model's training efficiency and shorten the training time, we plan to use transfer learning.

\section{Conclusion} \label{sec:Conclusion}

We propose WaferSegClassNet (WSCN), a lightweight deep learning architecture for performing the classification and segmentation of semiconductor wafer defects. We use N-pair contrastive loss to reduce the time to train and also boost classification and segmentation performance. 
The classification accuracy achieved by WSCN is higher than that of DCNet and ResNet50 and is in the same range as that of DenseNet121. The dice-score and IoU of WSCN are close to 1 and are competitive with open-source models such as DeepLabv3+ and UNet. WSCN's model size is 200$\times$ lower than that of ResNet50 and 250$\times$ lower than that of UNet. Thus, considering the balance of both predictive performance and model size, WSCN can be regarded as the best. We are the first to show segmentation performance on the MIXEDWM38 dataset. 
Our future work will extend our defect analysis technique to detect defects in other scenarios such as civil constructions.

\section*{Acknowledgement}
The experimental results were obtained using the computing resources provided by IIT Roorkee under the grant FIG-100874.

\bibliographystyle{IEEEtran}
\bibliography{References}
   
%\newpage


%\clearpage
%\newpage
%\input{Rough/ReviewerResponseRound3}

\end{document}